\pdfoutput=1

\documentclass[11pt]{article}

\usepackage{acl}
\usepackage{times}
\usepackage{latexsym}

\usepackage[T1]{fontenc}

\usepackage[utf8]{inputenc}

\usepackage{hyperref}       
\usepackage{url}            
\usepackage{booktabs}       
\usepackage{nicefrac}       
\usepackage{microtype}      
\usepackage{xcolor}         
\usepackage{graphicx}
\usepackage{amsmath}
\usepackage{amssymb}
\usepackage{mathtools}
\usepackage{amsthm}
\usepackage{array}
\usepackage{longtable}
\usepackage[most]{tcolorbox}
\usepackage{bibentry}
\usepackage{bm} 
\usepackage{tabularx}
\usepackage{multirow} 
\usepackage{algorithm}
\usepackage{algorithmic}
\usepackage{arydshln}
\usepackage{appendix}

\makeatletter
\def\blankfootnote{\xdef\@thefnmark{}\@footnotetext}
\makeatother

\definecolor{royalblue(web)}{rgb}{0.25, 0.41, 0.88}
\definecolor{blue-violet}{rgb}{0.54, 0.17, 0.89}
\definecolor{brightmaroon}{rgb}{0.76, 0.13, 0.28}
\definecolor{darkmagenta}{rgb}{0.55, 0.0, 0.55}
\definecolor{bleudefrance}{rgb}{0.19, 0.55, 0.91}
\definecolor{palatinateblue}{rgb}{0.15, 0.23, 0.89}
\definecolor{royalblue(web)}{rgb}{0.25, 0.41, 0.88}
\definecolor{whitesmoke}{rgb}{0.96, 0.96, 0.96}
\definecolor{thulianpink}{rgb}{0.87, 0.44, 0.63}
\definecolor{amber(sae/ece)}{rgb}{1.0, 0.49, 0.0}
\definecolor{darkblue}{rgb}{0.0, 0.0, 0.55}
\definecolor{alizarin}{rgb}{0.82, 0.1, 0.26}
\definecolor{asparagus}{rgb}{0.53, 0.66, 0.42}
\definecolor{darkspringgreen}{rgb}{0.09, 0.45, 0.27}
\definecolor{columbiablue}{rgb}{0.61, 0.87, 1.0}
\definecolor{wildblueyonder}{rgb}{0.64, 0.68, 0.82}
\definecolor{trolleygrey}{rgb}{0.5, 0.5, 0.5}
\definecolor{paleaqua}{rgb}{0.74, 0.83, 0.9}
\definecolor{bubblegum}{rgb}{0.99, 0.76, 0.8}
\definecolor{coralred}{rgb}{1.0, 0.25, 0.25}
\definecolor{green(ryb)}{rgb}{0.4, 0.69, 0.2}
\definecolor{flame}{rgb}{0.89, 0.35, 0.13}
\definecolor{bittersweet}{rgb}{1.0, 0.44, 0.37}
\definecolor{darksalmon}{rgb}{0.91, 0.59, 0.48}
\definecolor{emerald}{rgb}{0.31, 0.78, 0.47}
\definecolor{green(pigment)}{rgb}{0.0, 0.65, 0.31}
\definecolor{codegreen}{rgb}{0,0.6,0}
\definecolor{codegray}{rgb}{0.5,0.5,0.5}
\definecolor{codepurple}{rgb}{0.58,0,0.82}
\definecolor{backcolour}{rgb}{0.96,0.96,0.94}
\definecolor{bluegray}{rgb}{0.3, 0.38, 0.47}
\definecolor{whitesmoke}{rgb}{0.96, 0.96, 0.96}
\definecolor{codegreen}{rgb}{0,0.6,0}
\definecolor{codegray}{rgb}{0.5,0.5,0.5}
\definecolor{codepurple}{rgb}{0.58,0,0.82}
\definecolor{backcolour}{rgb}{0.96,0.96,0.94}

\tcbuselibrary{breakable}
\usepackage{listings}
\lstdefinestyle{mystyle}{
  basicstyle=\scriptsize\ttfamily,
  frame=single, 
  columns=fixed, 
}

\newtheorem{assumption}{Assumption}

\newtheorem{proposition}{Proposition}

\newcommand{\ours}{{\fontfamily{qpl}\selectfont PAFT}}

\usepackage{amsmath,amsfonts,bm}

\def\1{\bm{1}}

\DeclareMathAlphabet{\mathsfit}{\encodingdefault}{\sfdefault}{m}{sl}
\SetMathAlphabet{\mathsfit}{bold}{\encodingdefault}{\sfdefault}{bx}{n}
\newcommand{\tens}[1]{\bm{\mathsfit{#1}}}

\def\tI{{\tens{I}}}

\def\gR{{\mathcal{R}}}

\def\sD{{\mathbb{D}}}

\def\sP{{\mathbb{P}}}

\title{\ours{}: Prompt-Agnostic Fine-Tuning}

\author{
  Chenxing Wei$^{\dagger \S}$, Mingwen Ou$^{\circ}$, Ying He$^{\# \dagger}$, Yao Shu$^{\# \wr}$,  Fei Yu$^{ \ddagger}$\\
$^\dagger$College of Computer Science and Software Engineering, Shenzhen University, China \\
$^\circ$Tsinghua Shenzhen International Graduate School, Tsinghua University, China \\
$^{\S}$Guangdong Lab of AI and Digital Economy (SZ), China \\
$^{\wr}$Hong Kong University of Science and Technology (Guangzhou), China \\
$^{\ddagger}$School of Information Technology, Carleton University, Canada \\
\texttt{weichenxing2023@email.szu.edu.cn}, \texttt{yaoshu@hkust-gz.edu.cn}\\
}

\begin{document}
\maketitle
\begin{abstract}
Fine-tuning large language models (LLMs) often causes overfitting to specific prompt wording, where minor phrasing variations drastically reduce performance. To address this, we propose \textit{\underline{P}rompt-\underline{A}gnostic \underline{F}ine-\underline{T}uning} (\ours{}), a method that enhances robustness through dynamic prompt variation during training. \ours{} first generates diverse synthetic prompts, then continuously samples from this set to construct training instances, forcing models to learn fundamental task principles rather than surface-level patterns. Across systematic evaluations using both supervised fine-tuning (SFT) and reinforcement learning fine-tuning (RLFT), \ours{} demonstrates substantially improved prompt robustness, achieving 7\% higher generalization accuracy on unseen prompts than standard methods. In addition to enhanced robustness, \ours{} consistently yields superior overall performance on established benchmarks for question answering, mathematical reasoning, and tool use. Notably, models trained with \ours{} attain 3.2× faster inference speeds due to reduced prompt sensitivity. Ablation studies further validate  effectiveness of \ours{}, while theoretical analysis reveals that \ours{} can effectively enhance the cross-domain generalization ability of LLM.
\blankfootnote{${\#}$ corresponding author.}



\end{abstract}

\section{Introduction}
\label{introduction}
Large language models (LLMs) have demonstrated remarkable success across diverse natural language processing (NLP) tasks~\cite{zhao2024surveylargelanguagemodels, xu2023parameterefficientfinetuningmethodspretrained}. To further enhance the performance of LLMs on specific downstream tasks, supervised fine-tuning (SFT)~\cite{ouyang2022training, devlin-etal-2019-bert} and reinforcement learning fine-tuning (RLFT)~\cite{wang2024openchat, wei2025redit, wei2025tmpo} has emerged as a widely adopted strategy. These methods typically augment input data with task-specific instructions and construct dialogue datasets with expected outputs, enabling models to learn task-specific patterns. Empirical studies have shown that SFT and RLFT can substantially improve model performance on downstream tasks ~\cite{raffel2023exploringlimitstransferlearning, hu2023llmadapters, wei2022finetuned}. 
\begin{figure}[t]
\centering
\includegraphics[width=0.45\textwidth, trim={6cm 3cm 7cm 2cm}, clip]{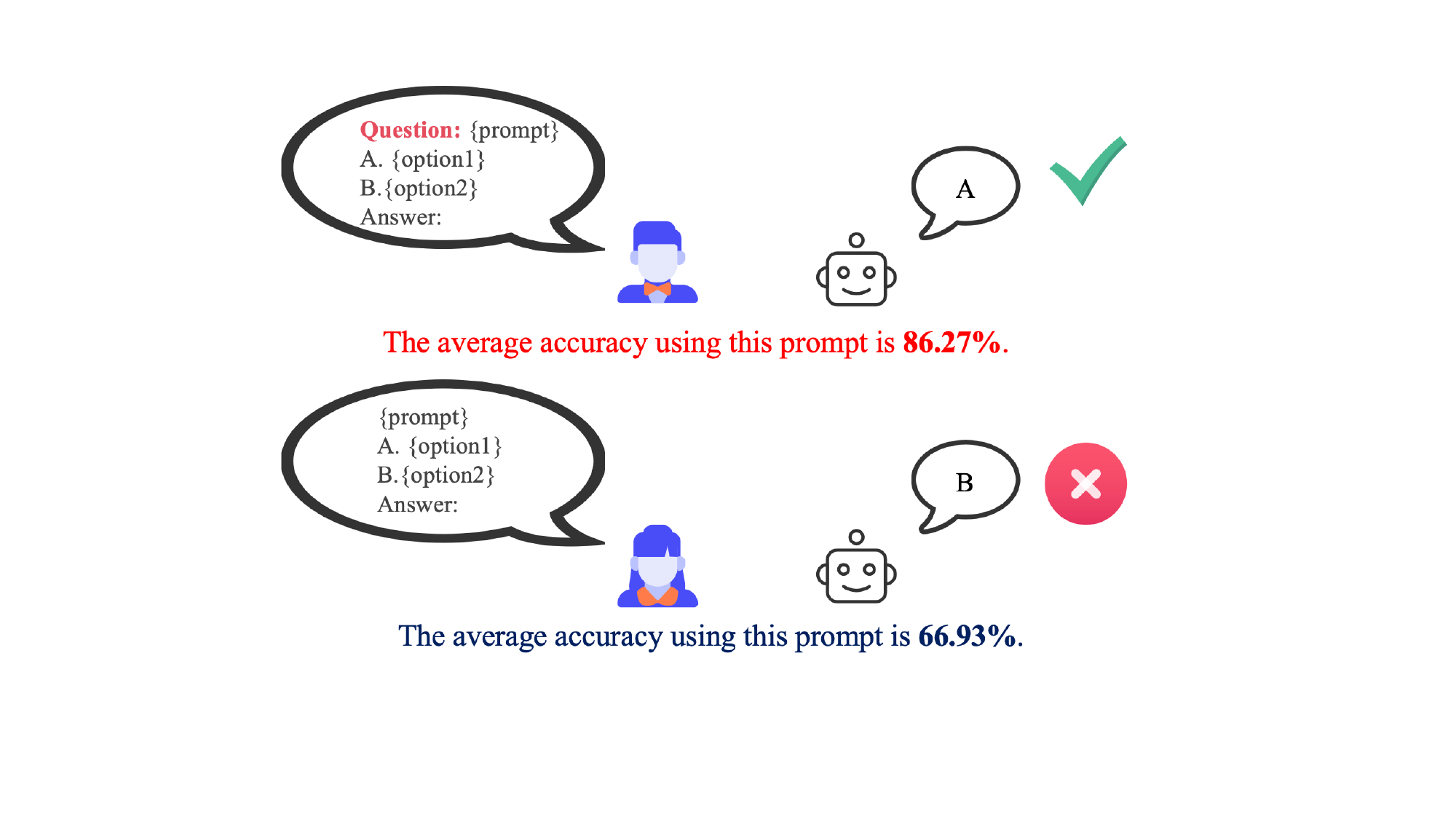}
\vspace{-2mm}
\caption{This figure shows how minor prompt changes drastically impact model accuracy. For instance, a one-word alteration to a prompt for the same user question reduced dataset accuracy from 86.27\% to 66.93\%. This highlights severe performance swings in models lacking prompt robustness.}
\label{fig:diglog}
\vspace{-3mm}
\end{figure}

\begin{figure*}[t]
\centering
\includegraphics[width=0.9\textwidth]{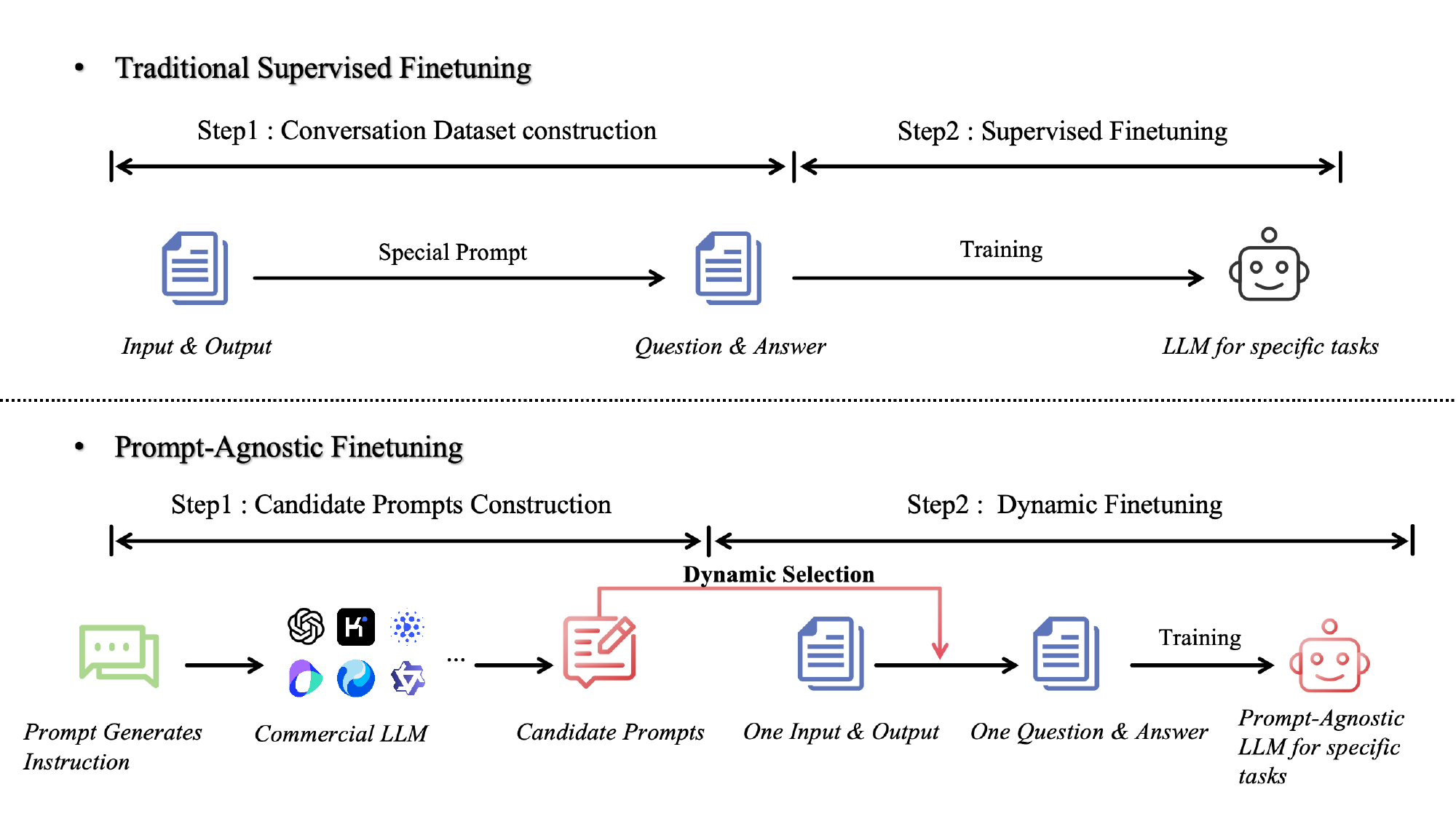}
\vspace{-5mm}
\caption{An overview of \ours{}: This figure contrasts SFT with \ours{}. While SFT relies on fixed datasets and predefined prompts—limiting robustness and cross-prompt generalization—\ours{} employs dynamic prompt selection during training, significantly enhancing prompt robustness and generalization capabilities. By leveraging commercial LLMs to generate diverse candidate prompts, \ours{} delivers a more scalable and generalizable solution for large language model adaptation.}
\label{fig:framework}
\vspace{-3mm}
\end{figure*}
However, as shown in Figure~\ref{fig:diglog}, a critical limitation of current fine-tuning methods is their lack of prompt robustness, as further detailed in Sec.~\ref{preliminaries}. Reliance on fixed instruction prompts \cite{mishra2022crosstaskgeneralizationnaturallanguage, chung2022scalinginstructionfinetunedlanguagemodels} often leads to overfitting on specific prompts patterns ~\cite{zhang2024instructiontuninglargelanguage, kung2023models}. Consequently, models become brittle: minor deviations between user and training prompts can significantly degrade inference performance~\cite{mialon2023augmented, raman2023modeltuning}. This brittleness manifests, for example, as substantial accuracy drops in QA tasks with altered prompt phrasing~\cite{wei2024flexora}, or as poor instruction following in chatbots and AI agents when commands deviate from those encountered during training~\cite{hong2024datainterpreterllmagent, sahoo2025systematicsurveypromptengineering}. Such sensitivity also raises fairness and reliability concerns in algorithmic comparisons~\cite{voronov-etal-2024-mind}. This vulnerability is particularly acute when users, unfamiliar with specific SFT prompt structures, provide highly divergent inputs, potentially causing fine-tuned models to perform near random guessing levels~\cite{polo2024efficient}. Notably, prompt robustness in SFT has received limited attention, with most existing work focusing on in-context learning and prompt tuning~\cite{ shi2024robustnessawareautomaticpromptoptimization,ishibashi-etal-2023-evaluating}.


To address this critical gap, we introduce \ours{}, a novel framework that dynamically adapts to diverse training prompts. To our knowledge, \ours{} is the first systematic approach to improving prompt robustness in both SFT and RLFT, a vital but underexplored area. Unlike traditional methods prone to overfitting specific prompt patterns, \ours{} enables models to grasp underlying task semantics, ensuring robust performance across varied human-written prompts. \ours{} operates in two phases (Figure~\ref{fig:framework}): first, constructing a diverse set of high-quality synthetic prompts that capture essential task semantics with linguistic variability (Sec.~\ref{sec:prompt}); second, employing dynamic fine-tuning by sampling from this curated set to expose the model to various formulations (Sec.~\ref{sec:finetuning}). Extensive evaluations demonstrate that \ours{} significantly boosts model robustness and generalization to diverse prompts, maintains state-of-the-art downstream performance, and can potentially improve inference speed while preserving training efficiency. These findings highlight \ours{} as a promising direction for developing more robust, user-friendly language models.

Our key contributions are as follows:
(a) We highlight that fine-tuning with fixed prompts results in poor generalization to unseen prompts and severe performance degradation (Sec.~\ref{preliminaries}).
(b) We propose \ours{}, a novel framework incorporating candidate prompt construction and dynamic fine-tuning, to enhance the prompt robustness of fine-tuned models (Sec.~\ref{framework}).
(c) We empirically demonstrate the consistent and robust performance of \ours{} across diverse downstream tasks, fine-tuning algorithms, and varied test prompts, including those unseen during training (Sec.~\ref{sec:results}).
(d) We provide theoretical evidence that \ours{} effectively enhances the cross-domain generalization of LLMs (Sec.~\ref{sec:theoretical_insights}).
\begin{figure*}[t]
\centering
\includegraphics[width=1.0\textwidth]{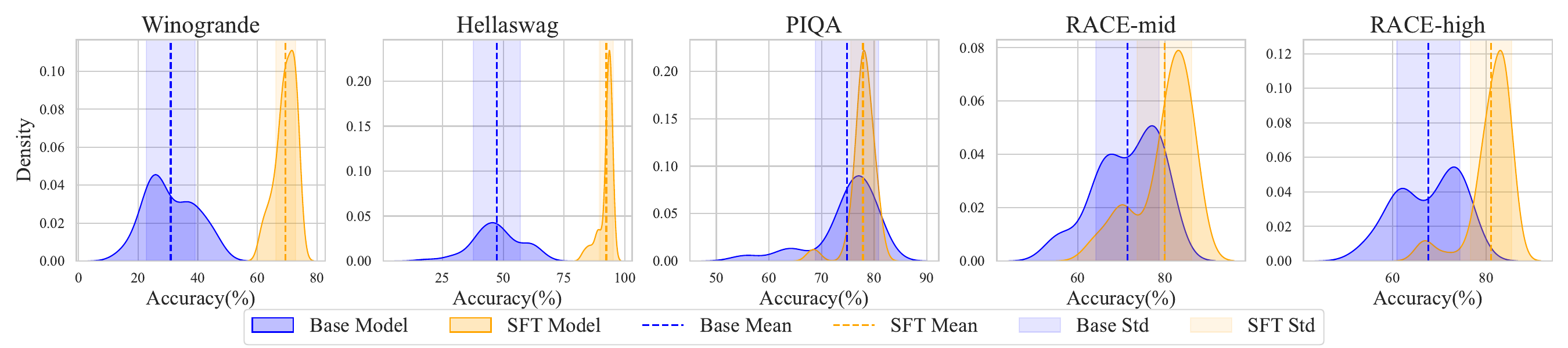}
\vspace{-5mm}
\caption{This figure presents experimental results across four datasets comparing base and SFT model performance on 450 diverse prompts (both human-written and LLM-generated). Probability distribution plots reveal that despite SFT's overall accuracy improvements, substantial performance variability persists—certain prompts yield markedly lower accuracy, with high standard deviations indicating significant prompt-dependent fluctuations. These findings underscore crucial impact of prompt and demonstrate the necessity for prompt-agnostic fine-tuning approaches.}
\label{fig:prompt_impact}
\vspace{-3mm}
\end{figure*}

\section{Related Work}
\label{sec:related_work}
\paragraph{Prompt Optimization.} Prompt engineering critically influences LLM performance, driving numerous prompt optimization approaches~\cite{chang2024efficient, li-2023-practical, diao2023blackbox, sun2022bbt}. Notable methods include INSTINCT~\cite{lin2024use}, which leverages neural network bandits with LLM embeddings for search efficiency. ZOPO~\cite{hu2024localized}, which employs localized search strategies. BATprompt~\cite{shi2024robustnessawareautomaticpromptoptimization}, which integrates robustness through natural language perturbations. While these approaches excel at identifying single high-performance prompts, models fine-tuned on such prompts remain vulnerable to prompt variations. Our work, in contrast, addresses this limitation by simultaneously enhancing prompt robustness and optimizing performance across the entire prompt space rather than focusing on isolated optimal prompts.

\paragraph{Fine-tuning (FT).} SFT and RLFT constitute the main paradigms for adapting LLMs to downstream tasks, prized for their efficiency. These approaches split into two categories: soft prompt tuning, which optimizes continuous input vectors while preserving base model parameters~\cite{kong2025meta, wu2024prompt, hu2024localized, lin2024use, li-liang-2021-prefix, liu-etal-2022-p}, and full/parameter efficient fine-tuning (PEFT)~\cite{shu2024ferret, ouyang2022training, liu2021pretrainpromptpredictsystematic, lester-etal-2021-power}. Among PEFT techniques, Low-Rank Adaptation (LoRA)~\cite{hu2022lora} predominates by freezing pre-trained weights while introducing trainable low-rank matrices, with recent variants enhancing generalization and reducing overfitting \cite{chen2023lorashearefficientlargelanguage, si2024unleashingpowertaskspecificdirections, wei2024flexora}. Instruction tuning \cite{sanh2022multitask} further improves ability of model to follow diverse task-specific instructions. However, existing methods—particularly soft prompt tuning—still exhibit limited prompt robustness, leaving models vulnerable to prompt variations. Our work addresses this critical limitation while maintaining computational efficiency.

\begin{figure*}[t]
\centering
\includegraphics[width=1.0\textwidth]{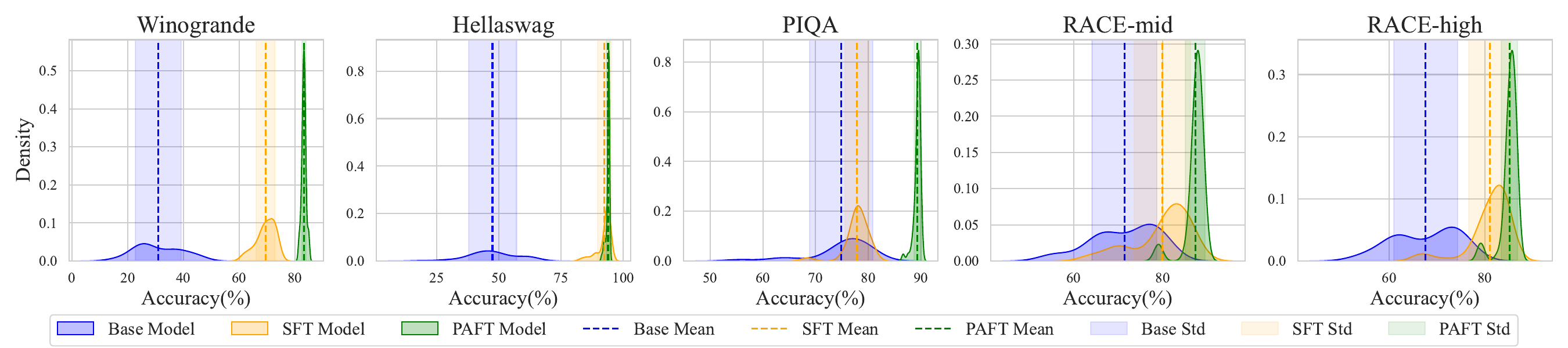}
\vspace{-5mm}
\caption{As a visual comparison to Figure~\ref{fig:prompt_impact}, we present performance distributions of base models, SFT models, and \ours{} across multiple reasoning and reading comprehension tasks. The probability distribution plots illustrate performance on unseen test prompts (both human-written and LLM-generated) not used during \ours{} training. Results clearly demonstrate \ours{} consistently achieves higher accuracy and lower variance across all tasks, confirming its effectiveness in enhancing prompt robustness.}
\label{fig:main_result}
\vspace{-3mm}
\end{figure*}

\section{Preliminaries}
\label{preliminaries}

To systematically study the impact of prompt variations on fine-tuned models, we conducted comprehensive experiments across multiple downstream tasks using LLaMA3-8B~\cite{llama3} with LoRA fine-tuning. We constructed a comprehensive set of over 450 prompts (both human-written and LLM-generated), covering a wide range of language styles, task-specific instructions, and formatting variations. Figure~\ref{fig:prompt_impact} presents a statistical analysis of the accuracy distribution for both the base and SFT models across these prompts, revealing a key finding: the formulation of the prompt dramatically influences the performance of the model regardless of the type of task, with only 10\% of the prompts producing near-optimal results. Minor prompt modifications (e.g., rephrasing, punctuation, reordering) induce substantial fluctuations. 

For example, the addition of "Question" improves accuracy by 20\% (Figure~\ref{fig:diglog}). This sensitivity highlights the fragility of current fine-tuning methods and their strong dependence on specific prompt formulations. These findings align with prior work~\cite{he2024doespromptformattingimpact, voronov-etal-2024-mind, salinas2024butterflyeffectalteringprompts, min-etal-2022-noisy, gao-etal-2021-making}. This widespread sensitivity demonstrates a fundamental limitation in current fine-tuning approaches, extending findings from previous research across diverse task domains. Based on these insights, we propose \ours{}, which decouples model performance from specific prompt formulations, ensuring consistent results across prompt variations and enhancing practical applicability in real-world scenarios.

\section{The \ours{} Framework}
\label{framework}
In this section, we introduce \ours{} in detail. As shown in Figure~\ref{fig:framework}, the \ours{} framework consists of two key stages: candidate prompt construction (Sec.~\ref{sec:prompt}) and dynamic fine-tuning (Sec.~\ref{sec:finetuning}).

\subsection{Candidate Prompt Construction}\label{sec:prompt}
To ensure the robustness and effectiveness of \ours{} across diverse prompts, we design a comprehensive prompt construction framework that aims to generate diverse and meaningful candidate prompts efficiently, enabling the model to generalize across different prompt formats. Our approach leverages the powerful generative capabilities of LLMs~\cite{kohl2024generativeaitoolkit} and comprises three key phases.

\textbf{Diverse LLM Ensemble.} We employ 10 mainstream LLMs with varied generation capabilities~\cite{openai2024gpt4technicalreport,bai2023qwentechnicalreport,ouyang2022training} to capture the inherent variability in task interpretation stemming from differences in pre-training data, architectures, and optimization objectives~\cite{minaee2024largelanguagemodelssurvey,zhao2024surveylargelanguagemodels}. This diversity ensures comprehensive coverage of prompt formulations across linguistic styles and instructional approaches, effectively mitigating single-model generation biases.

\textbf{Dual Prompting Strategy.} We combine \textbf{few-shot} and \textbf{zero-shot} techniques to balance quality and diversity. Few-shot prompting leverages in-context learning with curated human examples to generate task-aligned, semantically coherent prompts. Zero-shot prompting encourages diverse linguistic styles and structural variations without explicit examples. By generating 20 prompts with each strategy, we create a comprehensive set spanning high-quality and varied formulations, exposing the model to realistic prompt quality distributions and enhancing robustness to real-world scenarios. See Appendix~\ref{sec:prompt_generation_strategy} for details.

\textbf{Rigorous Evaluation Design.} We randomly partition generated prompts into training and test sets (8:1 ratio), ensuring completely distinct prompts in each set. This approach exposes the model to diverse prompt styles during training while providing a robust testbed for assessing generalization to novel formulations, see Appendix~\ref{app:prompt} for more details of prompt sets. By evaluating on entirely unseen prompts, we confirm that performance improvements reflect genuine ability to handle diverse prompt formulations rather than overfitting to specific patterns. This framework ensures \ours{} learns task semantics independently of prompt phrasing, enabling effective generalization across real-world scenarios while providing a scalable, cost-effective solution for improving prompt robustness.
\begin{algorithm}[t]
\caption{The \ours{} Framework}
\label{algo:paft}
\small
\begin{algorithmic}[1] 
    \STATE {\bfseries Input:} Generate a good candidate prompt training set $\sP$; A task-specific dataset $\sD$; The number of training epochs $T$; The number of same prompt training $K$; Initialized trainable parameters $\theta_0^0$; Learning rate $\eta_{\theta}$
    \STATE {\bfseries Output:} Fine-tuned model parameters $\theta^*$. 
    \FOR{each epoch $t=0$ {\bfseries to} $T-1$}
      \STATE $p \gets \text{RandomlySample}(\sP)$ \COMMENT{Randomly select a prompt from the candidate set}
      \STATE $k \gets 0$ \COMMENT{Initialize the step counter}
      \FOR{each data point $(x, y) \in \sD$}
        \STATE $\tI \gets \text{InputConstruction}(x, p)$ \COMMENT{Construct input using prompt $p$ and data $x$}
        \STATE $\theta_{t}^{k + 1} \gets \theta_{t}^k - \eta_{\theta} \nabla_{\theta} \ell(\theta, \tI)|_{\theta = \theta_{t}^k}$
        \STATE $k \gets k + 1$ \COMMENT{Increment the step counter}
        \IF{$k \mod K == 0$}
                \STATE $p \gets \text{RandomlySample}(\sP)$        
        \ENDIF
      \ENDFOR
      \STATE $\theta_{t+1}^0 \gets \theta_{t}^k$ 
    \ENDFOR
    \STATE {\bf return} $\theta^* = \theta_{T}^0$ 
\end{algorithmic}
\end{algorithm}
\begin{figure*}[t]
\centering
\includegraphics[width=1.0\textwidth]{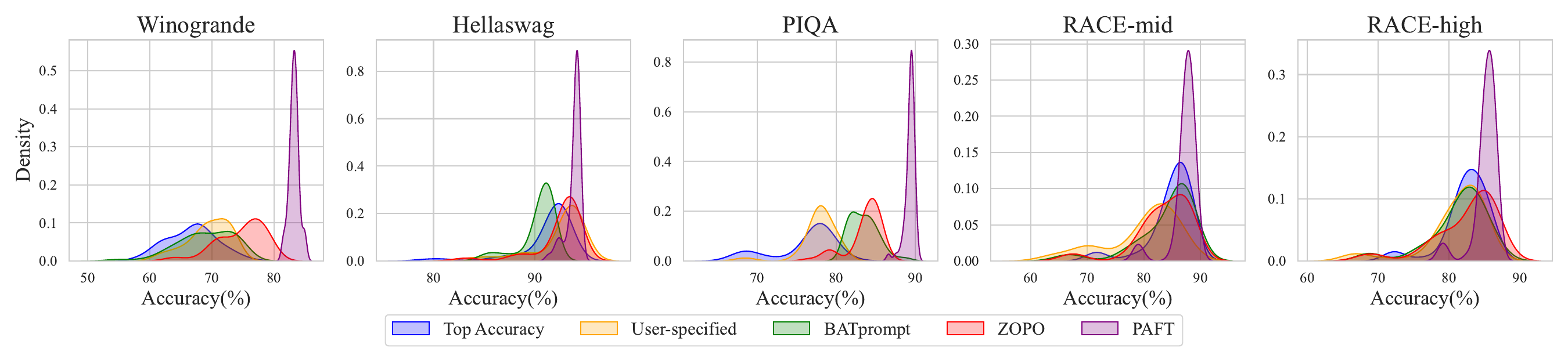}
\vspace{-5mm}
\caption{The performance of TopAccuracy, User-specified, BATprompt, ZOPO, and \ours{} models is compared on multiple reasoning and reading comprehension tasks. Results are reported in terms of their correct distribution. The tests are conducted on a test set of 50 unseen prompts, different from the ones used in training. The \ours{} model shows superior performance compared to other baselines, achieving higher accuracy and lower variance in all tasks.}
\label{fig:baseline}
\vspace{-3mm}
\end{figure*}
\subsection{Dynamic Fine-Tuning}\label{sec:finetuning}
\textbf{Dynamic Fine-Tuning Algorithm.} Our \ours{} framework enhances the robustness of LLMs through systematic prompt diversification. As shown in Algorithm~\ref{algo:paft}, each training epoch $t$ randomly samples a prompt $p$ from synthetic candidates $\sP$ (line 4), exposing the model to varied linguistic styles. For each data point $(x,y) \in \sD$ (line 6), the selected prompt is reused for $K$ consecutive steps (lines 7-9), constructing inputs via $\tI = \text{InputConstruction}(x, p)$ (line 7) and updating parameters $\theta$ using gradient-based optimization like SGD~\citep{sgd} or AdamW~\citep{adamw} (line 8). After $K$ steps, a new prompt is sampled (lines 10-11), ensuring multiple prompt exposures per epoch. Each epoch initializes with final parameters from the previous one: $\theta_{t+1}^0 = \theta_{t}^K$ (line 12), maintaining learning continuity until final parameters $\theta^* = \theta_{T}^0$ are achieved after $T$ epochs (line 16).

\textbf{Benefits of Dynamic Fine-Tuning.} Dynamic fine-tuning in \ours{} significantly enhances LLM robustness and generalization by exposing the model to diverse prompts during training. This approach mitigates overfitting to fixed prompts, fostering the learning of more generalizable representations less sensitive to specific formulations. Consequently, \ours{} achieves consistent performance across varied and unseen prompts, crucial for real-world applications with diverse user input. By reducing reliance on manual prompt engineering, dynamic fine-tuning offers an efficient and scalable solution for improving LLM adaptability.

\begin{table*}[t!]
    \centering
    \caption{Performance comparison of different fine-tuning methods on the test prompt sets across various reasoning and reading comprehension tasks using the LLaMA3-8B~\cite{llama3} with LoRA rank 8. Results are reported as average accuracy, standard deviation. \ours{} demonstrates superior performance, achieving the highest accuracy and lowest variance across all tasks. The last rows show the comparison of \ours{} with the second-best performing method (underlined). The Top column indicates the percentage of test prompts with a correct rate of 90\% for Hellaswag, 80\% for Winogrande, and 85\% for other datasets.}
    \label{tab:accuracy_comparison}
    \resizebox{\textwidth}{!}{
        \begin{tabular}{l|ccc|ccc|ccc|ccc|ccc|ccc}
            \toprule
            \textbf{Methods} & \multicolumn{3}{c|}{\textbf{Hellaswag}} & \multicolumn{3}{c|}{\textbf{PIQA}} & \multicolumn{3}{c|}{\textbf{Winogrande}} & \multicolumn{3}{c|}{\textbf{RACE-mid}}& \multicolumn{3}{c|}{\textbf{RACE-high}} & \multicolumn{3}{c}{\textbf{Average}} \\
            \midrule
            Metric & Mean & Std & Top & Mean & Std & Top & Mean & Std & Top & Mean & Std & Top & Mean & Std & Top & Mean & Std & Top \\
            \midrule
            Base Model & 47.36 & $\pm$9.78 & 0\% & 74.68 & $\pm$6.24 & 0\% & 45.15 & $\pm$11.78 & 0\% & 71.39 & $\pm$7.33 & 0\% & 67.62 & $\pm$6.78 & 0\% & 61.24 & $\pm$8.38 & 0\% \\
            User & 92.35 & $\pm$2.78 & 0\% & 77.87 & $\pm$2.36 & 0\% & \underline{78.16} & $\pm$7.97 & 0\% & 79.88 & $\pm$6.32 & 22\% & 81.05 & $\pm$4.45 & 4\% & 81.86 & $\pm$4.78 & 5\% \\
            TopAccuracy & 91.27 & $\pm$2.79 & \underline{86\%} & 75.96 & $\pm$3.89 & 0\% & 66.77 & $\pm$3.94 & 0\% & \underline{84.81} & \underline{$\pm$4.06} & 59\% & \underline{82.45} & \underline{$\pm$3.26} & 14\% & 80.25 & $\pm$3.63 & 32\% \\
            BATprompt & 90.30 & \underline{$\pm$1.79} & 78\% & 83.41 & \underline{$\pm$1.74} & 16\% & 69.01 & $\pm$4.45 & 0\% & 83.92 & $\pm$5.38 & \underline{65\%} & 81.33 & $\pm$4.21 & 12\% & 81.56 & \underline{$\pm$3.51} & 34\% \\
            ZOPO & \underline{92.46} & $\pm$2.43 & \underline{86\%} & \underline{83.52} & $\pm$2.23 & \underline{27\%} & 74.75 & \underline{$\pm$3.81} & 0\% & 83.50 & $\pm$5.05 & 51\% & 82.36 & $\pm$4.53 & \underline{35\%} & \underline{83.32} & $\pm$3.61 & \underline{40\%} \\
            \midrule
            \textbf{\ours{}} & \textbf{93.83} & $\pm$\textbf{0.70} & \textbf{100\%} & \textbf{89.33} & $\pm$\textbf{0.63} & \textbf{100\%} & \textbf{82.09} & $\pm$\textbf{0.81} & \textbf{100\%} & \textbf{87.26} & $\pm$\textbf{2.23} & \textbf{94\%} & \textbf{85.17} & $\pm$\textbf{1.71} & \textbf{73\%} & \textbf{87.57} & $\pm$\textbf{1.57} & \textbf{94\%} \\
            $\quad \hookrightarrow$ Improv. & +1.37 & -1.09 & 14\% & +5.81 & -1.11 & 73\% & +3.93 & -3.00 & 100\% & +2.45 & -1.83 & 29\% & +2.72 & -1.55 & 38\% & +4.25 & -1.94 & 54\% \\
            \bottomrule
        \end{tabular}
    }
    \vspace{-2mm}
\end{table*}

\begin{table*}[t!]
\centering
\caption{Experimental results on the HumanEval, Xstory\_cloze, Geometry3k, T-Eval, and GSM8K benchmarks. We compare our proposed \ours{} method with Base and SFT baselines. Performance is reported as Mean accuracy ($\pm$ Standard Deviation). The final row quantifies the absolute improvement of \ours{} over the standard SFT method for both mean and std. Best results from our method are highlighted in \textbf{bold}.}
\label{tab:exp_results_mean_std}
\resizebox{0.9\textwidth}{!}{
\begin{tabular}{@{}lccccc@{}}
\toprule
\textbf{Method} & \textbf{HumanEval} & \textbf{Xstory\_cloze} & \textbf{Geometry3k} & \textbf{T-Eval} & \textbf{GSM8K} \\
\midrule
Base & 41.31 ($\pm$ 10.36) & 48.23 ($\pm$ 8.36) & 32.17 ($\pm$ 15.36) & 58.97 ($\pm$ 14.03) & 74.36 ($\pm$ 21.37) \\
SFT  & 49.63 ($\pm$ 4.31) & 54.77 ($\pm$ 4.79) & 37.94 ($\pm$ 6.17) & 70.37 ($\pm$ 8.14) & 81.47 ($\pm$ 13.24) \\
\midrule
\textbf{\ours{}} & \textbf{54.24 ($\pm$ 1.36)} & \textbf{60.27 ($\pm$ 0.73)} & \textbf{40.19 ($\pm$ 1.27)} & \textbf{73.17 ($\pm$ 3.27)} & \textbf{85.71 ($\pm$ 5.93)} \\
\quad $\hookrightarrow$ Improv. & +4.61 (-2.95) & +5.50 (-4.06) & +2.25 (-4.90) & +2.80 (-4.87)& +4.24 (-7.31) \\
\bottomrule
\end{tabular}
}
\end{table*}

\begin{table}[t!]
    \centering
    \caption{Comparison of inference time (in hours) for different fine-tuning methods. \ours{} shows better inference efficiency than other methods. The last line shows the multiple of \ours{} improvement.}
    \label{tab:infer_time_comparison}
    \resizebox{0.48\textwidth}{!}{
        \begin{tabular}{l*{6}{c}}
            \toprule
            \textbf{Inference time/h} & \textbf{Hellaswag} & \textbf{PIQA} & \textbf{Winogrande} & \textbf{RACE} & \textbf{Average} \\
            \midrule
            Base Model & \underline{3.97} & 1.35 & \underline{1.72} & \underline{6.24} & \underline{3.32} \\
            User & 6.52 & 0.98 & 3.27 & 8.23 & 4.75 \\
            TopAccuracy  & 5.75 & 1.13 & 2.76 & 7.56 & 4.30 \\
            BATprompt & 4.57 & 1.57 & 3.14 & 7.98 & 4.32 \\
            ZOPO & 5.12 & \underline{0.87} & 3.23 & 8.28 & 4.38 \\
            \midrule
            \textbf{\ours{} } & \textbf{1.19} & \textbf{0.39} & \textbf{0.45} & \textbf{2.08} & \textbf{1.02} \\
            $\quad \hookrightarrow$ Improv. & $\times$3.3 & $\times$2.23 & $\times$3.82 & $\times$3.00 & $\times$3.25\\
            \bottomrule
        \end{tabular}
    }
    \vspace{-3mm}
\end{table}
\section{Empirical Results}\label{sec:results}
We evaluate our \ours{} framework through comprehensive experiments. Sec.~\ref{sec:setup} describes datasets and experimental setup, Sec.~\ref{sec:main result} analyzes key findings, and Sec.~\ref{ablation study} presents ablation studies examining critical framework components.

\subsection{Datasets and Setup}\label{sec:setup}

\textbf{Benchmark Selection.} 
As a pioneering work addressing prompt robustness in LLMs through training, we conduct experiments across a diverse set of tasks and benchmarks to ensure a comprehensive evaluation. Our methods involve Supervised Fine-Tuning (SFT) and a reinforcement learning approach, GRPO. For our SFT experiments, we selected benchmarks to cover a wide range of capabilities. Specifically, we use \textbf{HellaSwag}~\cite{HellaSwag} for knowledge understanding, \textbf{WinoGrande}~\cite{WinoGrande} for language understanding, and \textbf{RACE}~\cite{RACE} for reading reasoning capabilities. For grounding and abstractive summarization, we employ \textbf{PIQA}~\cite{PIQA}. To evaluate more specialized skills, we utilize \textbf{HumanEval}~\cite{chen2021codex} for coding, \textbf{T-Eval}~\cite{chen-etal-2024-eval} for tool use, and \textbf{Xstory\_cloze}~\cite{xstory_cloze} for multi-turn dialogues and multilingual tasks. For our GRPO experiments, we focus on mathematical reasoning. We use \textbf{GSM8K}~\cite{cobbe2021trainingverifierssolvemath} for math reasoning capabilities and \textbf{Geometry3k}~\cite{lu2021intergpsinterpretablegeometryproblem} for multimodal mathematical reasoning.

\textbf{Prompt Sets.} 
As detailed in Section~\ref{sec:prompt}, we constructed distinct prompt sets for training and evaluation. The training set contains 400 diverse prompts generated exclusively via LLMs; this approach demonstrates that \ours{} can fully automate the construction of training materials without manual intervention. For evaluation, we carefully designed a separate test set containing 50 prompts that includes not only LLM-generated prompts but also intentionally human-written instructions. The inclusion of human-written prompts is crucial for comprehensively validating the generalization of our model capabilities and its practical utility in real-world scenarios. This strict separation between a fully synthetic training set and a hybrid test set ensures a rigorous assessment of our method's effectiveness. Further details on the prompt generation process are provided in Appendix~\ref{app:prompt}.

\textbf{Baseline Comparisons.} We establish five baselines to isolate the impact of prompt engineering  on fine-tuning performance: the original pre-trained model (Base Model); the model fine-tuning with human-designed prompts (User) following~\citet{wei2024flexora}; the model fine-tuning with the highest accuracy of training prompts (TopAccuracy); the model fine-tuning with BATprompt~\cite{shi2024robustnessawareautomaticpromptoptimization}  most robust prompt (BATprompt); and fine-tuning with ZOPO~\cite{hu2024localized} optimal prompt selection (ZOPO). All models, including baselines, are evaluated on identical test prompts, enabling direct comparison of performance consistency across methods. 

\textbf{Experimental Setup.}
To comprehensively evaluate the effectiveness of \ours{}, our experiments cover two main training paradigms: supervised fine-tuning (SFT) and reinforcement learning fine-tuning (RLFT). For the SFT paradigm, we adopt Low-Rank Adaptation (LoRA~\cite{hu2022lora}) as a representative method, and the specific experimental parameters are shown in the Appendix~\ref{app:sft_setup}; for RLFT, we employ Group Relative Policy Optimization (GRPO~\cite{shao2024deepseekmathpushinglimitsmathematical}) and the specific experimental parameters are shown in the Appendix~\ref{app:grpo_setup}. In these settings, we utilize a series of large language models (LLMs), including Llama3-8B, Llama-3.1-8B, Llama-3.2-3B~\cite{llama3}, Qwen2.5-7B, and Qwen2.5-VL-7B~\cite{qwen2025qwen25technicalreport}. Detailed correspondence between model datasets and training paradigms is provided in Appendix Table~\ref{tab:task_distribution}. Our implementation is based on the Llama-factory framework, and all evaluations are performed using OpenCompass. All experiments are conducted on NVIDIA A100, V100, 4090, and L40 GPU clusters. For detailed configuration, see the Appendix~\ref{app:setup}.

\subsection{Main Results}\label{sec:main result}

\begin{figure}[t!]
\centering
\includegraphics[width=0.4\textwidth]{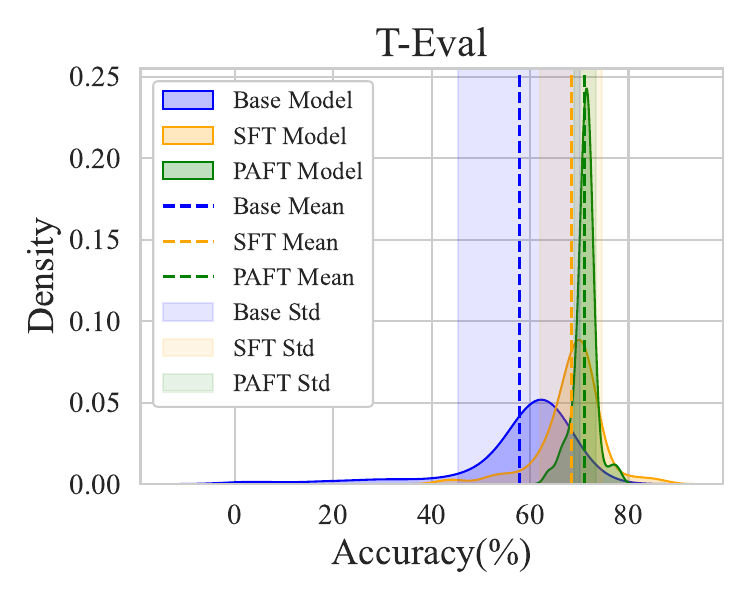}
\caption{The performance of base model, SFT model, and \ours{} model is compared on T-Eval.}
\label{fig:teval_accuracy}
\vspace{-3mm}
\end{figure}

\begin{figure}[t!]
\centering
\includegraphics[width=0.4\textwidth]{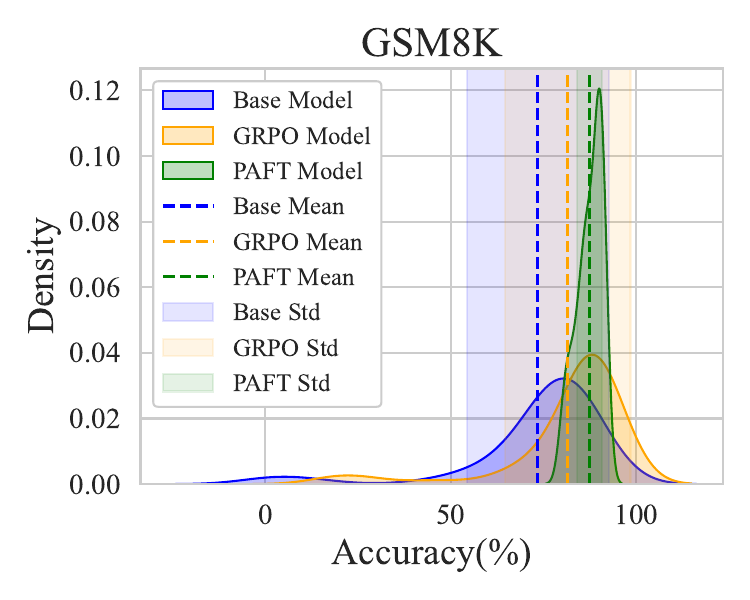}
\caption{The performance of base model, GRPO model, and \ours{} model is compared on GSM8K.}
\label{fig:gsm8k_accuracy}
\vspace{-3mm}
\end{figure}

\textbf{Prompt Robustness.} As demonstrated across Tables~\ref{tab:accuracy_comparison}, Table~\ref{tab:exp_results_mean_std}, Figures~\ref{fig:main_result}, \ref{fig:baseline}, \ref{fig:teval_accuracy}, and \ref{fig:gsm8k_accuracy}, \ours{} exhibits remarkably low variance across all evaluation tasks, indicating superior prompt robustness. This enhanced stability stems from our dynamic prompt selection strategy (Sec.~\ref{sec:finetuning}), which continuously adjusts prompts during training, compelling the model to learn essential task features rather than overfitting to specific prompt formats. In contrast, baseline approaches face significant limitations: user prompts rely on manual design with inconsistent quality; TopAccuracy and ZOPO tend to overfit to high-performing training prompts with poor generalization; and while BATprompt addresses robustness, it remains less effective than our method. The low variance of \ours{} translates to more stable performance and stronger generalization across diverse prompts, enabling development of more user-friendly QA systems, format-independent agent systems, and directly evaluate the true ability of LLMs by better decoupling the ability from the prompting engineering. Notably, \ours{} achieves acceptable performance across most prompts, significantly outperforming all baselines (Table~\ref{tab:accuracy_comparison}, Top column) while maintaining high training efficiency (detailed in Appendix~\ref{app:train time}).

\begin{table}[t!]
\centering
\small 
\caption{Comparison of Minimum and Conditional Accuracy (\%). Min. Acc. is on 50 unseen prompts; Cond. Acc. is on 10 adversarial prompts.}
\label{tab:min_cond_accuracy} 
\resizebox{0.45\textwidth}{!}{
\begin{tabular}{@{}lrrrrrr@{}} 
\toprule
& \multicolumn{2}{c}{\textbf{SFT Model}} & \multicolumn{2}{c}{\textbf{\ours{} Model}} & \multicolumn{2}{c}{\textbf{Improvement}} \\
\cmidrule(lr){2-3} \cmidrule(lr){4-5} \cmidrule(lr){6-7} 
\textbf{Dataset} & \textbf{Min} & \textbf{Con} & \textbf{Min} & \textbf{Con} & \textbf{Min} & \textbf{Con} \\
\midrule
HellaSwag   & 87.20 & 61.26 & \textbf{91.30} & \textbf{84.61} & +4.10  & +23.35 \\
PIQA        & 75.16 & 62.13 & \textbf{88.72} & \textbf{84.12} & +13.56 & +22.99 \\
HumanEval   & 45.26 & 12.46 & \textbf{52.89} & \textbf{50.16} & +7.63  & +37.70 \\
RACE-mid    & 72.36 & 50.16 & \textbf{85.07} & \textbf{83.27} & +12.71 & +33.11 \\
RACE-high   & 71.68 & 49.67 & \textbf{84.26} & \textbf{81.26} & +12.58 & +31.59 \\
GSM8K       & 40.36 & 10.26 & \textbf{75.13} & \textbf{74.13} & +34.77 & +63.87 \\
Geometry3k  & 31.26 & 12.37 & \textbf{38.90} & \textbf{37.12} & +7.64  & +24.75 \\
T-Eval      & 42.13 & 19.26 & \textbf{61.27} & \textbf{59.17} & +19.14 & +39.91 \\

\bottomrule
\end{tabular}
}
\end{table}

To quantify this robustness under more demanding conditions, we introduce two stringent metrics. First, we measure \textbf{minimum accuracy (Min)} on the test set of 50 unseen prompts to evaluate worst-case performance under normal conditions. Second, we assess \textbf{conditional accuracy (Con)} on a challenging set of 10 adversarially crafted prompts containing paraphrases, misspellings, or other modifications to measure resilience against noisy inputs. As shown in Table~\ref{tab:min_cond_accuracy}, the \ours{}-trained model achieves significantly higher minimum and conditional accuracy across all datasets. 
This result underscores the ability of \ours{} to maintain effective performance even when faced with substantial prompt perturbations and adversarial noise.

\textbf{SOTA Performance.} 
As demonstrated across our experiments (Tables~\ref{tab:accuracy_comparison} and~\ref{tab:exp_results_mean_std}; Figures~\ref{fig:main_result}--\ref{fig:gsm8k_accuracy}), \ours{} consistently achieves SOTA performance by significantly outperforming existing baselines. This superior performance stems directly from our Dynamic Fine-Tuning algorithm (Algorithm~\ref{algo:paft}; Section~\ref{sec:finetuning}), which effectively decouples the underlying fundamental principles of a task from any specific prompt formulation. This decoupling allows the model to focus on learning the essential features of task, rather than becoming entangled in the nuances of a particular prompt. Ultimately, this process enables the model to learn the fundamental principles of downstream tasks instead of merely overfitting to superficial prompt patterns, which is the key reason why \ours{} model can achieve its SOTA generalization and performance.

\begin{table*}[t!]
    \centering
    \caption{Performance comparison of \ours{} with varying hyperparameters \(K\) (number of iterations per prompt) and \(T\) (number of epochs) across multiple reasoning and reading comprehension tasks. Results are reported as mean accuracy (\(\pm\) standard deviation) on the Hellaswag, PIQA, Winogrande, RACE-mid, and RACE-high datasets. The best results for each metric are highlighted in bold.}
    \label{tab:kt_comparison}
    \resizebox{\textwidth}{!}{
        \begin{tabular}{l*{6}{c}}
            \toprule
            \textbf{\# \( K \) and \( T \)} & \textbf{Hellaswag} & \textbf{PIQA} & \textbf{Winogrande} & \textbf{RACE-mid}& \textbf{RACE-high} & \textbf{Average} \\
            \midrule
            \( K \) = 1, \( T \) = 3 & 93.58 ($\pm$ 1.47) & \textbf{89.33} ($\pm$ 0.63) & 81.78 ($\pm$ 1.11) & 86.30 ($\pm$ 2.73) & 84.35 ($\pm$ 2.24) & 87.07 ($\pm$ 1.64) \\
            \( K \) = 2, \( T \) = 3 & 93.59 ($\pm$ 1.24) & 88.37 ($\pm$ \textbf{0.49}) & \textbf{82.09} ($\pm$ \textbf{0.81}) & 86.30 ($\pm$ 2.64) & 84.02 ($\pm$ 2.24) & 86.87 ($\pm$ 1.48) \\
            \( K \) = 4, \( T \) = 3 & \textbf{93.83} ($\pm$ 1.10) & 89.07 ($\pm$ 0.53) & 81.96 ($\pm$ 1.15) & \textbf{87.26} ($\pm$ \textbf{2.23}) & \textbf{85.17} ($\pm$ 1.71) & \textbf{87.46} ($\pm$ \textbf{1.34}) \\
            \( K \) = 8, \( T \) = 3 & \textbf{93.83} ($\pm$ \textbf{0.70}) & 88.99 ($\pm$ 0.59) & 82.69 ($\pm$ 0.97) & 86.25 ($\pm$ 2.75) & 84.36 ($\pm$ 2.06) & 87.22 ($\pm$ 1.41) \\
            \( K \) = 1, \( T \) = 6 & 93.37 ($\pm$ 1.47) & 88.32 ($\pm$ 0.68) & 81.05 ($\pm$ 3.44) & 84.40 ($\pm$ 2.30) & 83.34 ($\pm$ \textbf{1.66}) & 86.10 ($\pm$ 1.91) \\
            \bottomrule
        \end{tabular}
    }
    \vspace{-3mm}
\end{table*}

\begin{figure*}[t]
\centering
\includegraphics[width=1.0\textwidth]{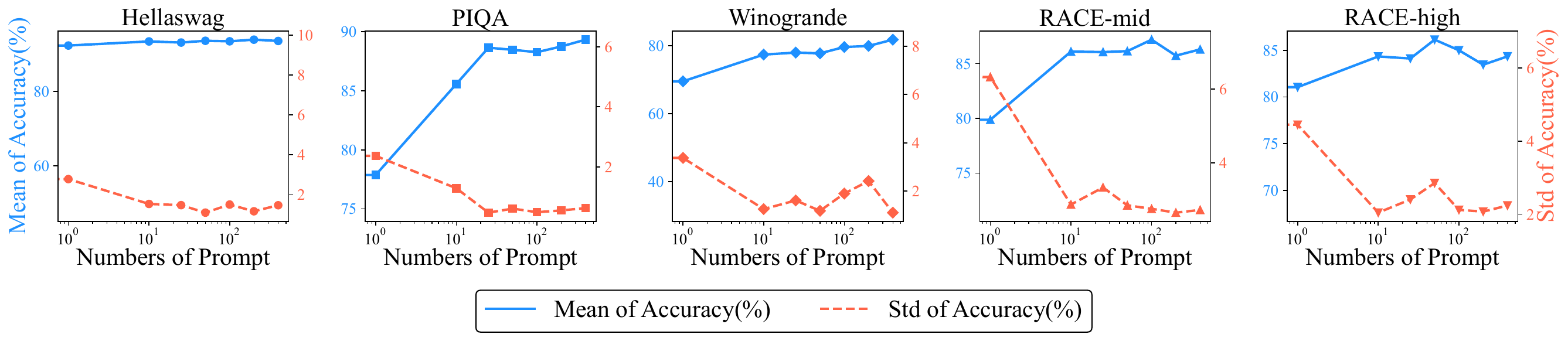}
\caption{Scaling Law of Training Prompt Numbers: Mean and Standard Deviation of Accuracy Across Different Datasets. The x-axis represents the number of prompts on a logarithmic scale, while the y-axis shows the mean accuracy (left) and standard deviation of accuracy (right) for each dataset.}
\label{fig:number of prompts}
\vspace{-3mm}
\end{figure*}

\textbf{Inference Efficiency.} 
\ours{} enhances inference efficiency not by accelerating per-token generation speed, but by enabling the model to produce correct and concise responses using significantly fewer tokens. Our measurements across all test prompts and datasets (Table~\ref{tab:infer_time_comparison}) demonstrate that this token reduction leads to consistently faster overall inference times compared to baseline methods. This efficiency stems from the prompt robustness of model, which is a core outcome of our training approach. Unlike baseline models that may overfit to superficial prompt patterns, \ours{} learns underlying task principles. As a result, even in the face of changing or modified prompts, the model remains focused on the fundamental task objectives, avoiding the need to output redundant dialogue content, such as understanding unknown instructions. This making \ours{} particularly valuable for real-world applications requiring rapid responses, such as dialogue systems and AI agents, while simultaneously reducing computational resource requirements. See Appendix~\ref{app:infer} for more detail.

\subsection{Ablation Studies} \label{ablation study}
\textbf{Hyperparameter Robustness.} This ablation study demonstrates the robustness of \ours{} to the hyperparameters \(K\) (iterations per prompt) and \(T\) (epochs). As shown in Table~\ref{tab:kt_comparison}, \ours{} achieves stable performance across a broad range of \(K\) (1 to 8) and \(T\) (3 to 6) values, with minimal fluctuations in accuracy and variance. Notably, \ours{} achieves near-optimal performance with default settings (\(K=4\), \(T=3\)), attaining an average accuracy of 87.46\%(\(\pm 1.34\)) across all tasks. This robustness reduces the need for extensive hyperparameter tuning, making \ours{} a practical and efficient solution for real-world applications.

\textbf{Impact of Training Prompt Quantity.}
We conduct an ablation study to investigate the impact of varying numbers of training prompts on model performance, thus validating the effectiveness of \ours{}. The experimental results, shown in Figure~\ref{fig:number of prompts}, demonstrate that as the number of prompts increases, the average accuracy of the model significantly improves, while the standard deviation decreases, indicating more stable and reliable performance. However, the performance gains diminish as the number of prompts increases, with only marginal improvements observed beyond a certain threshold. This suggests that while adding prompts can enhance performance, \ours{} achieves competitive results with a minimal number of prompts, rendering excessive prompts unnecessary. In most cases, \ours{} achieves strong performance with as few as 10 high-quality prompts, and further increases yield only marginal gains. The efficiency of \ours{} is particularly notable, as it delivers excellent performance with a minimal number of prompts, making it highly suitable for resource-constrained scenarios where computational efficiency is critical. These findings underscore the practicality and efficiency of \ours{}, offering a robust and efficient solution for real-world applications.

\section{Theoretical Insights}
\label{sec:theoretical_insights}

The capability of \ours{} to generalize effectively to unseen prompt formulations can be rigorously understood through the lens of domain adaptation theory~\cite{ben2006analysis, ben2010theory}. In this theoretical construct, the collection of training prompts $\mathcal{P}_{\text{train}}$ along with the task-specific training data $\mathcal{D}_{\text{train}}$ delineates the source domain. Besdies, the set of novel test prompts $\mathcal{P}_{\text{test}}$, paired with $\mathcal{D}_{\text{test}}$, represents the target domain. \ours{} aims to learn a model $f^* \in \mathcal{H}$, where $\mathcal{H}$ denotes the hypothesis class, by minimizing the empirical risk computed over instances $(x, p_i, y)$ where each prompt $p_i$ is sampled from $\mathcal{P}_{\text{train}}$.

A foundational result from domain adaptation theory~\cite{ben2010theory} provides an upper bound on the expected risk of $f^*$ on the target prompt distribution $\gR_{\mathcal{P}_{\text{test}}}(f^*)$ with $\min_{f \in \mathcal{H}} (\gR_{\mathcal{P}_{\text{train}}}(f) + \gR_{\mathcal{P}_{\text{test}}}(f))$:
\begin{equation}
\begin{split}
    \gR_{\mathcal{P}_{\text{test}}}(f^*) & \le  \text{Disc}(\mathcal{P}_{\text{train}}, \mathcal{P}_{\text{test}})\\
    & \quad + \mathcal{C}(\mathcal{H}, N) + \hat{\gR}_{\mathcal{P}_{\text{train}}, N}(f^*) + \lambda^* \ .
\end{split}
\end{equation}

Here, $\hat{\gR}_{\mathcal{P}_{\text{train}}, N}(f^*)$ is the empirical risk on $N$ training prompts. The term $\mathcal{C}(\mathcal{H}, N)$ signifies model complexity (e.g., related to Rademacher complexity~\cite{yin2020rademachercomplexityadversariallyrobust}), which typically diminishes as the number of distinct training prompts  $N$ increase; this term captures the generalization gap on the source domain. The divergence between the training and test prompt distributions is quantified by $\text{Disc}(\mathcal{P}_{\text{train}}, \mathcal{P}_{\text{test}})$. Finally, $\lambda^*$ encapsulates the optimal joint error achievable by a hypothesis in $\mathcal{H}$ on both domains. The key of \ours{} is designed to optimize this bound for improved generalization.

\textbf{Complexity Control.} By employing a substantial number of distinct training prompts $N$, \ours{} inherently works to reduce the complexity term $\mathcal{C}(\mathcal{H}, N)$. This ensures that the model performance observed on the training prompts becomes a more faithful estimator of its true performance across the entire $\mathcal{P}_{\text{train}}$ distribution, fostering more stable learning. This effect is empirically supported by our ablation studies in Section~\ref{ablation study} (Figure~\ref{fig:number of prompts}), which demonstrate improved stability with more prompts. 

\begin{figure}[t]
\centering
\includegraphics[width=0.4\textwidth]{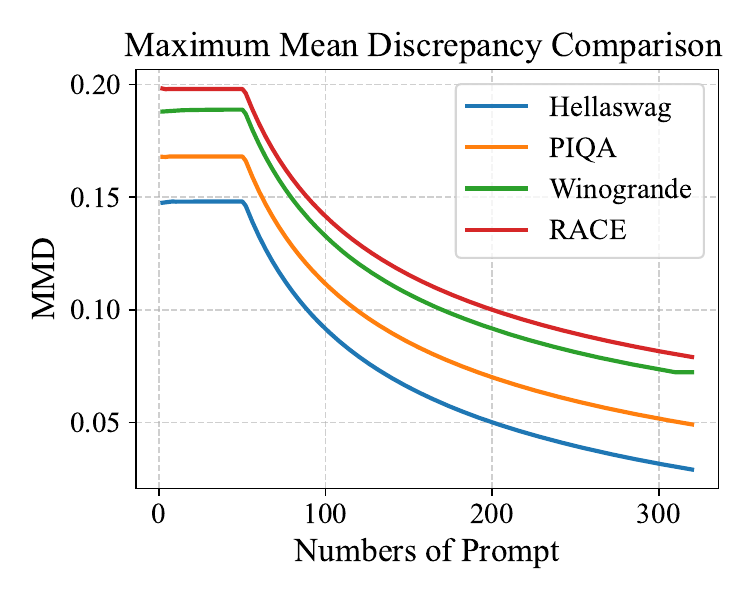}
\caption{This figure demonstrates the change in MMD for different numbers of $\mathcal{P}_{\text{train}}$ and the same $\mathcal{P}_{\text{test}}$.}
\label{fig:mmd}
\vspace{-3mm}
\end{figure}

\textbf{Domain Alignment.} 
Minimizing the domain discrepancy term $\text{Disc}(\mathcal{P}_{\text{train}}, \mathcal{P}_{\text{test}})$ is critically dependent on constructing a diverse and comprehensive set of candidate prompts $\mathcal{P}_{\text{train}}$ (see Section \ref{sec:prompt} for details). A more diverse $\mathcal{P}_{\text{train}}$ is more likely to effectively cover or closely approximate the diverse and unseen distribution of test prompts $\mathcal{P}_{\text{test}}$. This approximation reduces the divergence between the training and test prompt distributions and enhances the transferability of knowledge learned from $\mathcal{P}_{\text{train}}$ to $\mathcal{P}_{\text{test}}$. 

\begin{proposition}[MMD as an Upper Bound on Discrepancy]
\label{prop:mmd_bound_combined}
The discrepancy term $\text{Disc}(\mathcal{P}, \mathcal{Q})$ can be bounded by the Maximum Mean Discrepancy (MMD)~\cite{gao2021maximummeandiscrepancytest} upper bound:
$$ \text{Disc}(\mathcal{P}, \mathcal{Q}) \le C \cdot \text{MMD}(\mathcal{P}, \mathcal{Q}) $$
where MMD is defined as:
$$ \text{MMD}(\mathcal{P}, \mathcal{Q}) = \sup_{g \in \mathcal{H}, \|g\|_{\mathcal{H}} \le 1} \left| \mathbb{E}_{\mathcal{P}}[g] - \mathbb{E}_{\mathcal{Q}}[g] \right| $$
\end{proposition}
The proof of Proposition~\ref{prop:mmd_bound_combined} is provided in the Appendix~\ref{sec:proof}. Therefore, we can quantify the upper bound of domain difference using MMD. As illustrated in Figure~\ref{fig:mmd}, an increasing number of diverse training prompts cover a wider semantic space, bringing $\mathcal{P}_{\text{train}}$ closer to $\mathcal{P}_{\text{test}}$. This proximity reduces the upper bound of the target prompt distribution.
 
\textbf{Generalization Guarantee.} By minimizing the empirical risk $\hat{R}_{\mathcal{P}_{\text{train}}, N}(f^*)$ across a sufficiently large and varied corpus of prompts, \ours{} encourages the model to internalize the underlying task semantics, rather than merely memorizing superficial prompt structures. This principled approach is key to improving the model performance $R_{\mathcal{P}_{\text{test}}}(f^*)$ when confronted with novel and unencountered prompts.

\section{Conclusion}

\ours{} offers a compelling solution for enhancing the prompt robustness of LLMs. By dynamically adjusting prompts during fine-tuning, \ours{} significantly improves model prompt robustness and performance across diverse prompt formulations. Notably, \ours{} boosts inference speed with maintained training cost. This approach paves the way for more reliable and efficient LLM deployment in real-world applications.

\newpage

\section*{Acknowledgement}
This work is supported in part by Shenzhen Science and Technology Program under Grant ZDSYS20220527171400002, the National Natural Science Foundation of China (NSFC) under Grants 62271324, 62231020 and 62371309.

\section*{Limitations}
In this section, we discuss potential limitations of \ours{} and outline promising directions for future research. While \ours{} demonstrates significant progress in enhancing the prompt robustness of Large Language Models (LLMs), certain aspects warrant further investigation. A key area for improvement lies in the dynamic prompt selection strategy employed during fine-tuning.  Currently, \ours{} utilizes a random sampling approach, which, while exposing the model to a diverse range of prompts, may not be the most efficient or effective method.  Exploring more sophisticated sampling techniques, such as curriculum learning or importance sampling, could potentially optimize the training process and further enhance robustness. For instance, prioritizing prompts that induce higher loss or those that are more representative of the overall prompt distribution could lead to faster convergence and improved generalization. Furthermore, integrating adversarial learning into the dynamic fine-tuning phase presents a compelling avenue for future work. Generating adversarial prompts on-the-fly, perhaps through gradient-based updates, could further challenge the model and encourage it to learn more robust task representations. This approach could be particularly beneficial in mitigating the impact of maliciously crafted or unexpected prompts. However, the well-known instability of adversarial training remains a significant hurdle.  Stabilizing the training process, perhaps through techniques like robust optimization or regularization, is crucial for realizing the full potential of this approach.  Investigating different adversarial prompt generation strategies and their impact on model robustness would be a valuable contribution.

\section*{Ethics Statement}
We have manually reevaluated the dataset we created to ensure it is free of any potential for discrimination, human rights violations, bias, exploitation, and any other ethical concerns.
\newpage

\bibliography{anthology}

\appendix
\onecolumn

\section{Theoretical Proof}
\label{sec:proof}
We first state a core assumption regarding the richness of the reproducing kernel Hilbert space (RKHS).

\begin{assumption}[Richness of RKHS]
\label{assump:rkhs_richness}
We assume that the RKHS $\mathcal{H}$ generated by the chosen kernel function $k$ is large enough to contain all 1-Lipschitz functions. Formally, we assume that there exists a constant $C > 0$ such that any function $f$ with a Lipschitz constant at most 1 (i.e., $|f|_{\text{Lip}} \le 1$) is contained in $\mathcal{H}$ and has a bounded norm, i.e., $\|f\|_{\mathcal{H}} \le C$.
\end{assumption}

\begin{proof}[Proof Sketch]
The difference term is the supremum over all 1-Lipschitz functions. According to the assumption ~\ref{assump:rkhs_richness}, we can extend this function space to the RKHS sphere of radius $C$, which contains all 1-Lipschitz functions, and then obtain the definition of MMD by rescaling.
\begin{align*} 
\text{Disc}(\mathcal{P}, \mathcal{Q}) &= \sup_{f: |f|_{\text{Lip}} \le 1} \left| \mathbb{E}_{\mathcal{P}}[f] - \mathbb{E}_{\mathcal{Q}}[f] \right| \\ 
&\le \sup_{g \in \mathcal{H}: \|g\|_{\mathcal{H}} \le C} \left| \mathbb{E}_{\mathcal{P}}[g] - \mathbb{E}_{\mathcal{Q}}[g] \right| \\ 
&= C \cdot \sup_{h \in \mathcal{H}: \|h\|_{\mathcal{H}} \le 1} \left| \mathbb{E}_{\mathcal{P}}[h] - \mathbb{E}_{\mathcal{Q}}[h] \right| \\
&= C \cdot \text{MMD}(\mathcal{P}, \mathcal{Q})
\end{align*}
This derivation formally shows that we can use MMD to quantize $\text{Disc}(\mathcal{P}, \mathcal{Q}) $.
\end{proof}

\section{Experimental setting}\label{app:setup}
In the main experiment, we compared \ours{} with the baseline. The datasets and experimental parameters are as follows:
\subsection{Dataset}

In this section, we introduce the statistics of the dataset. The statistics of the dataset are shown in Table~\ref{tab:Statistics For Data}. 

\begin{table*}[h]
\centering
\caption{Number of samples in the train, validation, and test datasets for various dateset.}
\label{tab:Statistics For Data}
\begin{tabular}{l*{4}{c}}
\toprule
\textbf{Number of samples} &\textbf{train dataset} & \textbf{validation dataset} & \textbf{test dataset} \\ 
\midrule
Hellaswag & 39900 & 10000 & 10000  \\
 PIQA & 16000 & 2000 & 3000  \\
Winogrande & 40398 & 1267 & 1767  \\
 RACE & 87866 & 4887 & 4934  \\
\bottomrule
\end{tabular}
\end{table*}

\begin{table}[h!]
\centering
\caption{Task Distribution Across Datasets}
\label{tab:task_distribution}
\begin{tabular}{@{}llll@{}}
\toprule
\textbf{Type} & \textbf{Method} & \textbf{LLM} & \textbf{Dataset} \\
\midrule
Knowledge Understanding & SFT & Llama3-8B & HellaSwag \\
Language Understanding & SFT & Llama3-8B & WinoGrande \\
Math Reasoning Capabilities & GRPO & Qwen2.5-7B & GSM8K \\
Reading Reasoning Capabilities & SFT & Llama3-8B & RACE \\
Grounding and Abstractive Summarization & SFT & Llama3-8B & PIQA \\
Coding Capabilities & SFT & Qwen2.5-7B & HumanEval \\
Tool use & SFT & Llama-3.1-8B & T-Eval \\
Multi-turn dialogues and multilingual tasks & SFT & Llama-3.2-3B & Xstory\_cloze \\
Multimodal mathematical reasoning & GRPO & Qwen2.5-VL-7B & Geometry3k \\
\bottomrule
\end{tabular}
\end{table}

\subsection{Specific SFT experimental parameters} 
\label{app:sft_setup}
Based on the LLaMA3-8B model configuration, several adjustments were made to optimize model performance. In the baseline model experiment, generation parameters were adjusted to ensure the correct output. In the LoRA experiment, adjustments to the generation parameters were retained, and LoRA-related parameters were adjusted. In the \ours{} experiment, the size of the validation set was adjusted to control the time required to search for the optimal layer. For specific experimental parameters, see the table~\ref{tab:parameters}.

\begin{table*}[h]
\centering
\caption{Detailed experimental parameters. This table lists the specific parameters we used in the experiments for various methods. These parameters include the target module of LoRA (Lora Target), the maximum sequence length (Max Length), the number of samples for supervised fine-tuning (SFT Samples), the learning rate (LR), the number of training prompts (Training Prompts). Epoch(Epoch) represents the epoch of training. All other parameters not listed here remain consistent across all experiments. }
\label{tab:parameters}
\begin{tabular}{l*{7}{c}}
\toprule
\textbf{Methods} &\textbf{LoRA Target} & \textbf{Max Length} & \textbf{SFT Samples} & \textbf{LR} & \textbf{Training Prompts} & \textbf{Epoch} \\ 
\midrule
LoRA & q \& v Proj & 1024 & 20000 & 0.0001 & 1 & 3 \\
\midrule
\ours{} & q \& v Proj & 1024 & 20000 & 0.0001 & 400 & 3\\
\bottomrule
\end{tabular}
\end{table*}

\subsection{\ours{} Integration with GRPO}
\label{app:grpo_setup}

To demonstrate the versatility of our \ours{} framework beyond SFT, we also integrated it with the Reinforcement Learning Fine-Tuning (RLFT) paradigm. We selected Group Relative Policy Optimization (GRPO) as a representative RLFT method and applied our dynamic prompting strategy to its training process for the GSM8K dataset. The setup is as follows:

\textbf{Candidate Prompt Construction.}
Following the core principle of \ours{}, we first constructed a diverse candidate prompt set for mathematical reasoning. We utilized multiple large language models (LLMs) to generate a wide variety of prompts related to math problems, which formed the prompt training set. To ensure a rigorous evaluation, additional prompts from real dialogues and other synthetic sources were used to create a distinct test set, guaranteeing that the prompt training and test sets were entirely distinct.

\textbf{Dynamic GRPO Training Process.}
We integrated our dynamic sampling mechanism directly into the GRPO training loop. In each step of the standard GRPO process, instead of using a fixed instruction, we randomly sampled a prompt from our candidate training set. This sampled prompt (e.g., \textit{``Please help me solve this math problem \{GSM8K\_problem\}''}) was then combined with a problem instance from the GSM8K dataset to form the final input for the model's computation step. This process aligns with the standard procedure for Soft Fine-Tuning.

\textbf{Hyperparameters.}
The key hyperparameters used for our GRPO experiments are detailed in Table~\ref{tab:grpo_params}.

\begin{table}[h!]
\centering
\caption{GRPO Hyperparameters.}
\label{tab:grpo_params}
\begin{tabular}{@{}lc@{}}
\toprule
\textbf{Parameter} & \textbf{Value} \\
\midrule
Learning Rate & 5e-6  \\
Num Generations & 16  \\
Epochs & 10  \\
\bottomrule
\end{tabular}
\end{table}

\section{Training cost and inference time}\label{app:train time}
\subsection{Training cost}
\textbf{\ours{} Maintains Training Efficiency.} We now turn our attention to the training efficiency of \ours{}. A critical consideration for any practical fine-tuning approach is its impact on training time.  Introducing complex mechanisms or additional computational overhead can significantly hinder the training process, especially when dealing with large language models and extensive datasets.  Therefore, it is essential to demonstrate that \ours{} does not introduce such burdens.

To rigorously evaluate the training time implications of \ours{}, we conducted a series of experiments, using Low-Rank Adaptation (LoRA) \cite{hu2022lora} as a representative example of a parameter-efficient fine-tuning method. LoRA has gained popularity due to its ability to adapt pre-trained models with minimal computational cost, making it a suitable baseline for our analysis.  Our experiments, the results of which are presented in Table~\ref{tab:time_comparison}, directly compare the training time required for traditional LoRA fine-tuning with the training time required for \ours{} integrated with LoRA.

The key finding from our analysis is that \ours{} does not introduce any noticeable increase in training time.  The data in Table~\ref{tab:time_comparison} clearly demonstrates that the training duration remains virtually identical whether we employ standard LoRA or incorporate \ours{}'s dynamic prompt selection mechanism. This crucial observation underscores the efficiency of \ours{}.  The dynamic prompt selection process, which is central to \ours{}'s ability to enhance prompt robustness, is implemented in a way that does not add significant computational overhead.  This is because the selection process is lightweight and seamlessly integrated into the existing training loop.  Rather than requiring complex computations or extensive data manipulations, \ours{} efficiently chooses from a diverse set of prompts, allowing the model to experience a wider range of input formulations without incurring a substantial time penalty.  This efficient dynamic prompt selection is critical for the practical applicability of \ours{}, ensuring that it can be readily deployed without compromising training efficiency.  Furthermore, this efficiency allows for more extensive experimentation and exploration of different prompt variations, ultimately leading to more robust and generalizable models.

\begin{table*}[t!]
    \centering
    \caption{Training Time Comparison of Different Fine-tuning Methods on the Test Prompt Sets Across Various Reasoning and Reading Comprehension Tasks Using the LLaMA3-8B\cite{llama3} Model with LoRA Rank 8. Experiments were conducted on an NVIDIA RTX 4090 GPU. Results are reported as training time in hours. \textbf{LoRA + TopAccuracy prompt } prompt refers to the prompt with the highest accuracy in the training set, \textbf{LoRA + user-specified prompt }\cite{wei2024flexora} refers to fine-tuning with human-designed prompts, \textbf{LoRA + BATprompt} \cite{shi2024robustnessawareautomaticpromptoptimization} uses the most robust prompt generated by BATprompt, and \textbf{LoRA + ZOPO prompt} \cite{hu2024localized} employs the optimal prompt selected by ZOPO from the training prompt set.}
     
    \label{tab:time_comparison}
    \resizebox{\textwidth}{!}{
        \begin{tabular}{l*{6}{c}}
            \toprule
            \textbf{Training time/h} & \textbf{Hellaswag} & \textbf{PIQA} & \textbf{Winogrande} & \textbf{RACE} & \textbf{Average} \\
            \midrule
            LoRA + user-specified prompt & 3.01 & 2.35 & 3.27 & 3.95 & 3.15 \\
            LoRA + TopAccuracy prompt & 3.00 & 2.29 & 2.98 & 3.93 & 3.05 \\
            LoRA + BATprompt & 3.02 & \textbf{2.23} & 3 & 3.93 & 3.05 \\
            LoRA + ZOPO prompt & \textbf{2.97} & 2.3 & \textbf{2.97} & 3.83 & \textbf{3.02} \\
            \midrule
            \textbf{\ours{} } & 2.98 & 2.32 & 3.38 & \textbf{3.81} & 3.12 \\
            \bottomrule
        \end{tabular}
    }
    \vspace{-3mm}
\end{table*}

\textbf{Efficient Candidate Prompt Generation.}  A key aspect of \ours{}'s effectiveness lies in its ability to generate a diverse and high-quality set of candidate prompts efficiently.  The process of constructing these candidate prompts involves leveraging the capabilities of external large language models (LLMs), which naturally raises the question of associated costs.  Specifically, we sought to quantify the token usage required for candidate prompt generation, as this directly translates to the expense incurred when interacting with commercial LLM APIs.

To address this, we conducted a detailed analysis of the token consumption during the candidate prompt generation phase of \ours{}.  Our investigation, the results of which are summarized in Table~\ref{tab:tokens}, focuses on the number of tokens required to produce a sufficient variety of prompts suitable for subsequent selection and fine-tuning.  We meticulously tracked the token usage across various prompts generated for different tasks, considering factors such as prompt length, complexity, and diversity.

The findings presented in Table~\ref{tab:tokens} demonstrate that \ours{} requires remarkably few tokens to generate a substantial pool of candidate prompts.  This efficiency stems from \ours{}'s strategic approach to prompt engineering.  Rather than relying on brute-force generation or computationally intensive search methods, \ours{} employs a carefully designed prompting strategy that encourages the external LLMs to produce a wide range of prompt formulations with minimal token consumption.  This is achieved through techniques such as few-shot prompting with carefully chosen examples,  targeted instructions that guide the LLM towards desired prompt characteristics, and potentially iterative refinement of prompts based on preliminary evaluation.  The low token count is crucial for practical applications, as it minimizes the cost associated with using commercial LLM APIs.  Moreover, this efficiency enables the exploration of a broader range of potential prompts within a fixed budget, increasing the likelihood of discovering highly effective prompts that contribute to improved model robustness.  This efficient prompt generation process is a significant advantage of \ours{}, enabling it to achieve superior performance without incurring prohibitive costs.

\begin{table*}[t!]
    \centering
    \scriptsize
    \caption{Token Usage for Candidate Prompt Generation. This table shows the number of tokens used to generate approximately 400 candidate prompts for each task. The average token usage is 11.75k. The number of generated prompts can be adjusted based on the scaling law observed in Figure~\ref{fig:number of prompts} to control costs.}
    \label{tab:tokens}
    \resizebox{\textwidth}{!}{
        \begin{tabular}{l*{6}{c}}
            \toprule
            \textbf{Tokens} & \textbf{Hellaswag} & \textbf{PIQA} & \textbf{Winogrande} & \textbf{RACE} & \textbf{Average} \\
            \midrule
            Total Tokens & 11.7k & 12.1k & 10.9k & 12.3k & 11.75k \\
            \bottomrule
        \end{tabular}
    }
    \vspace{-3mm}
\end{table*}
\subsection{Inference Times}
In this section, we show some special cases to demonstrate that \ours{} improves reasoning efficiency by improving the model’s understanding of the core task semantics, enabling concise and accurate responses with fewer tokens.
\begin{tcolorbox}[title={\textbf{\small cases study}},
colback=whitesmoke, colframe=darkblue, , boxrule=2pt, arc=0mm]
{\scriptsize
\begin{lstlisting}[style=mystyle]
Hellaswage: {
    "prompt": "By analyzing the setup in A man is sitting on a roof.
    He, deduce which outcome would serve as the best fit: 
    A. is using wrap to wrap a pair of skis.\n
    B. is ripping level tiles off.\n
    C. is holding a rubik's cube.\n
    D. starts pulling up roofing on a roof.\n 
    You may choose from 'A', 'B', 'C', 'D'.\n Answer:",
    "SFT prediction": "Best answer: D, as "pulling up roofing on a roof" is the most plausible action 
    in this context, while the others are either irrelevant or unclear.",
    "PAFT prediction": "D",
    "gold": "D"
},
PIQA: {
    "origin_prompt": "For the situation described by ice box, consider which solution aligns more
    closely with how things usually work in real life: 
    A. will turn into a cooler if you add water to it\n
    B. will turn into a cooler if you add soda to it. 
    Use logical reasoning to guide your choice. Answer:",
    "SFT prediciton": "The correct answer is A. will turn into a cooler if you add water to it, 
    because: A traditional ice box (or cooler) works by using ice (frozen water) to keep items cold.
    Adding water to an ice box would help maintain cooling as the ice melts,
    whereas soda (a sugary, non-freezing liquid) would not improve cooling efficiency.",
    "PAFT prediciton" : "A",
    "gold": "A"
},
Winograde: {
    "origin_prompt": "
    In the context of Sarah was a much better surgeon than 
    Maria so _ always got the harder cases.\n
    which word best completes the sentence? Choose:
    A. Sarah\n
    B. Maria\n
    Answer:",
    "SFT prediciton": "Maybe Sarah or Maria I might need more information to answer this question. 
    I guess the final answer is B.",
    "PAFT prediciton" : "A",
    "gold": "A"
},
\end{lstlisting}
}
\end{tcolorbox}
\label{app:infer}
\section{Prompt}\label{app:prompt}
In this section, we present a selection of training and test prompts to illustrate the efficacy of our prompt construction algorithm and to provide a clearer understanding of operational process of \ours{}. Due to space constraints, we only list 10 prompts as examples. Section~\ref{sec:prompt_generation_strategy} shows how we guide LLMs to generate candidate prompts. Section~\ref{train prompt} showcases examples of training prompts, Section~\ref{test prompt} highlights test prompts, and Section~\ref{baseline prompt} outlines the prompts utilized by the baseline method.

\subsection{Automated Prompt Generation Strategy}
\label{sec:prompt_generation_strategy}

To construct a diverse and high-quality set of candidate prompts, we employ a strategy that leverages large language models (LLMs) through two distinct templating approaches: zero-shot and one-shot prompting. These templates are designed to be general, requiring only minor modifications to synthesize prompt sets for various datasets.

\textbf{Zero-shot Prompting.} Our zero-shot approach uses a general template that instructs an LLM to generate multiple prompt variations for a given task type, without being constrained by a specific problem instance. This method is effective for tasks where the output format is straightforward. For example, to generate prompts for a commonsense reasoning task like PIQA, we use the following instruction:

\begin{tcolorbox}[title={\textbf{\small Train Prompt of Hellaswag}},
colback=whitesmoke, colframe=darkblue, , boxrule=2pt, arc=0mm]
{\scriptsize
\begin{lstlisting}[style=mystyle]
Please write 20 detailed English prompts for me to solve a commonsense reasoning problem... 
You don't need to design a specific problem, just design a template, and replace the problem
description with a question. Requirements: diverse styles, lengths, and structures.
\end{lstlisting}
}
\end{tcolorbox}

\textbf{One-shot Prompting.} For tasks requiring a specific output format, such as the step-by-step reasoning in mathematical problems, we use a one-shot template. This template provides the LLM with an explicit example of the desired output structure in addition to the generation instructions, thereby guiding the model's response format[cite: 1]. For datasets like GSM8K and Geometry3K, our one-shot instruction is as follows:

\begin{tcolorbox}[title={\textbf{\small Train Prompt of Hellaswag}},
colback=whitesmoke, colframe=darkblue, , boxrule=2pt, arc=0mm]
{\scriptsize
\begin{lstlisting}[style=mystyle]
Please write 20 detailed English prompts for me to solve a math problem... 
An example: Here is the question: \{question\}, 
let's think step by step and respond in the following format: 
<reasoning>...</reasoning><answer>...</answer>
\end{lstlisting}
}
\end{tcolorbox}

This templating strategy is designed to produce prompts that are both adaptable and general. By crafting instructions that elicit the necessary information without being overly task-specific, we ensure the generated prompts can be applied across different datasets with minimal modification. This approach is fundamental to our goal of enhancing prompt robustness and practical applicability, demonstrating that our framework can automate the creation of effective and varied training prompts.

\subsection{Train prompt}\label{train prompt}
In this section, we present the prompts generated using the method outlined in Section~\ref{sec:prompt} across various datasets. All prompts listed here are utilized for training purposes.
\begin{tcolorbox}[title={\textbf{\small Train Prompt of Hellaswag}},
colback=whitesmoke, colframe=darkblue, , boxrule=2pt, arc=0mm]
{\scriptsize
\begin{lstlisting}[style=mystyle]
1. Based on the given context {ctx}, which of the following options correctly predicts the outcome?
Choose the correct letter option.\n A. {A}\nB. {B}\nC. {C}\nD. {D}\n Answer:
2. Considering the scenario described in {ctx}, identify the most accurate prediction of the 
final result:Select the correct letter.\n A. {A}\nB. {B}\nC. {C}\nD. {D}\n Answer:
3. Given the information in {ctx}, which option best forecasts the correct ending?Provide the 
correct letter choice.\n A. {A}\nB. {B}\nC. {C}\nD. {D}\n Answer:
4. From the context {ctx}, which of the following options accurately predicts the conclusion?Write
down the correct letter.\n A. {A}\nB. {B}\nC. {C}\nD. {D}\n Answer:
5. Using the details provided in {ctx}, select the option that correctly predicts the final outcome:
Enter the correct letter.\n A. {A}\nB. {B}\nC. {C}\nD. {D}\n Answer:
6. Based on the context {ctx}, which option is the most accurate prediction of the ending?Choose the
correct letter option.\n A. {A}\nB. {B}\nC. {C}\nD. {D}\n Answer:
7. Given the scenario in {ctx}, identify the option that correctly forecasts the outcome:Select the
correct letter.\n A. {A}\nB. {B}\nC. {C}\nD. {D}\n Answer:
8. Considering the details in {ctx}, which option best predicts the correct conclusion?Provide the
correct letter choice.\n A. {A}\nB. {B}\nC. {C}\nD. {D}\n Answer:
9.Analyze the context {ctx} and determine the correct prediction of the outcome:Indicate the 
correct letter.\n A. {A}\nB. {B}\nC. {C}\nD. {D}\n Answer:
10. Analyze the given context {ctx} and determine the most accurate prediction of the final result:
Indicate the correct letter.\n A. {A}\nB. {B}\nC. {C}\nD. {D}\n Answer:
\end{lstlisting}
}
\end{tcolorbox}
\begin{tcolorbox}[title={\textbf{\small Train Prompt of PIQA}},
colback=whitesmoke, colframe=darkblue, , boxrule=2pt, arc=0mm]
{\scriptsize
\begin{lstlisting}[style=mystyle]
1.In order to {goal}, which of the following options is the most logical choice based on common
knowledge?\nA. {sol1}\nB. {sol2}\nAnswer:
2.Consider the scenario where you need to {goal}. Which option would be the most appropriate 
according to general understanding?\nA. {sol1}\nB. {sol2}\nAnswer:
3.When trying to {goal}, which of the following would be the best course of action based on everyday
reasoning?\nA. {sol1}\nB. {sol2}\nAnswer:
4.To achieve {goal}, which option aligns best with common sense?\nA. {sol1}\nB. {sol2}\nAnswer:
5.Based on typical knowledge, which of the following is the correct choice to {goal}?
\nA. {sol1}\nB. {sol2}\nAnswer:
6.If you want to {goal}, which of these options would be the most sensible according to common
reasoning?\nA. {sol1}\nB. {sol2}\nAnswer:
7.Using general knowledge, determine the best option to {goal}.\nA. {sol1}\nB. {sol2}\nAnswer:
8.To {goal}, which of the following choices is the most reasonable based on common sense?
\nA. {sol1}\nB. {sol2}\nAnswer:
9.When considering how to {goal}, which option would be the most logical based on everyday knowledge?
\nA. {sol1}\nB. {sol2}\nAnswer:
10.According to common reasoning, which of the following is the best way to {goal}?
\nA. {sol1}\nB. {sol2}\nAnswer:
\end{lstlisting}
}
\end{tcolorbox}
\begin{tcolorbox}[title={\textbf{\small Train Prompt of Winogrande}},
colback=whitesmoke, colframe=darkblue, , boxrule=2pt, arc=0mm]
{\scriptsize
\begin{lstlisting}[style=mystyle]
1.Choose the correct answer to complete the sentence.{ctx}
\nA. {only_option1}\nB. {only_option2}\nAnswer:
2.elect the appropriate option to fill in the blank.{ctx}
\nA. {only_option1}\nB. {only_option2}\nAnswer:
3.Fill in the blank with the correct answer.{ctx}
\nA. {only_option1}\nB. {only_option2}\nAnswer:
4.Identify the correct choice to complete the statement.{ctx}
\nA. {only_option1}\nB. {only_option2}\nAnswer:
5.Choose the right answer to fill in the gap .{ctx}
\nA. {only_option1}\nB. {only_option2}\nAnswer:
6.Select the correct option to complete the sentence.{ctx}
\nA. {only_option1}\nB. {only_option2}\nAnswer:
7.Fill in the blank with the correct answer.{ctx}
\nA. {only_option1}\nB. {only_option2}\nAnswer:
8.Identify the correct choice to complete the sentence.{ctx}
\nA. {only_option1}\nB. {only_option2}\nAnswer:
9.Choose the right answer to fill in the blank. {ctx}
\nA. {only_option1}\nB. {only_option2}\nAnswer:
10.Select the appropriate option to complete the statement.{ctx}
\nA. {only_option1}\nB. {only_option2}\nAnswer:
\end{lstlisting}
}
\end{tcolorbox}
\begin{tcolorbox}[title={\textbf{\small Train Prompt of RACE}},
colback=whitesmoke, colframe=darkblue, , boxrule=2pt, arc=0mm]
{\scriptsize
\begin{lstlisting}[style=mystyle]
1.Carefully read the following article and answer the question by selecting the correct option.
Respond with A, B, C, or D.\n\nArticle:\n{article}\n\n
Q: {question}\n\nA. {A}\nB. {B}\nC. {C}\nD. {D}\nAnswer:
2.Read the passage below and choose the best answer to the question.
Reply with the letter A, B, C, or D.\n\nArticle:\n{article}\n\n
Q: {question}\n\nA. {A}\nB. {B}\nC. {C}\nD. {D}\nAnswer:
3.After reading the article, answer the following question by selecting the correct option.
Please respond with A, B, C, or D.\n\nArticle:\n{article}\n\n
Q: {question}\n\nA. {A}\nB. {B}\nC. {C}\nD. {D}\nAnswer:
4.Examine the article provided and answer the question by choosing the most appropriate option.
Reply with A, B, C, or D.\n\nArticle:\n{article}\n\n
Q: {question}\n\nA. {A}\nB. {B}\nC. {C}\nD. {D}\nAnswer:
5.Read the following text and answer the question by selecting the correct letter.
Respond with A, B, C, or D.\n\nArticle:\n{article}\n\n
Q: {question}\n\nA. {A}\nB. {B}\nC. {C}\nD. {D}\nAnswer:
6.Carefully read the article and choose the best answer to the question.
Reply with A, B, C, or D.\n\nArticle:\n{article}\n\n
Q: {question}\n\nA. {A}\nB. {B}\nC. {C}\nD. {D}\nAnswer:
7.Read the passage and answer the question by selecting the correct option.
Respond with A, B, C, or D.\n\nArticle:\n{article}\n\n
Q: {question}\n\nA. {A}\nB. {B}\nC. {C}\nD. {D}\nAnswer:
8.After reading the article, choose the correct answer to the question.
Reply with A, B, C, or D.\n\nArticle:\n{article}\n\n
Q: {question}\n\nA. {A}\nB. {B}\nC. {C}\nD. {D}\nAnswer:
9.Read the provided text and answer the question by selecting the best option.
Respond with A, B, C, or D.\n\nArticle:\n{article}\n\n
Q: {question}\n\nA. {A}\nB. {B}\nC. {C}\nD. {D}\nAnswer:
10.Examine the article and answer the question by choosing the correct letter.
zReply with A, B, C, or D.\n\nArticle:\n{article}\n\n
Q: {question}\n\nA. {A}\nB. {B}\nC. {C}\nD. {D}\nAnswer:
\end{lstlisting}
}
\end{tcolorbox}
\subsection{Test prompt}\label{test prompt}
In this section, we present the prompts generated using the method outlined in Section~\ref{sec:prompt} across various datasets. All prompts listed here are utilized for testing purposes, and they are not visible during training.
\begin{tcolorbox}[title={\textbf{\small Test Prompt of Hellaswag}},
colback=whitesmoke, colframe=darkblue, , boxrule=2pt, arc=0mm]
{\scriptsize
\begin{lstlisting}[style=mystyle]
1.Based on the information provided, please select the most probable conclusion: {ctx}
\n A. {A}\nB. {B}\nC. {C}\nD. {D}\n 
Remember to consider the implications of each option. Answer:
2.In the scenario described by {ctx}, there is only one correct way the story or situation could end.
When predicting the right ending, consider the cause-and-effect relationships established within 
the context.An option that logically follows from the preceding events is likely the correct one.
\n A. {A}\nB. {B}\nC. {C}\nD. {D}\n You may choose from 'A', 'B', 'C', 'D'.\n Answer:
3.Based on the given context {ctx}, which of the following options correctly predicts the outcome?
Choose the correct letter option.
\n A. {A}\nB. {B}\nC. {C}\nD. {D}\n Answer:
4.To solve this problem based on {ctx}, weigh the significance of each potential ending:
A. {A}\nB. {B}\nC. {C}\nD. {D}\n You may choose from 'A', 'B', 'C', 'D'.\n Answer:
5.Analyzing the context of {ctx}, think about the relationships and conflicts presented.
Which option is most likely to resolve these issues and lead to a satisfying ending?
\n A. {A}\nB. {B}\nC. {C}\nD. {D}\n Answer:
6.{ctx}\nQuestion: Taking into account the context, which outcome is the most expected?
\n A. {A}\nB. {B}\nC. {C}\nD. {D}\n Answer:
7.From the detailed description provided, choose the option that best completes the scenario:{ctx}\
n A. {A}\nB. {B}\nC. {C}\nD. {D}\n 
Consider all aspects of the scenario to make an informed decision on the correct ending.\n Answer:
8.Given the scenario described in {ctx}, which of the following conclusions seems most plausible? 
Consider all the details and clues provided to make an informed guess.
\n A. {A}\nB. {B}\nC. {C}\nD. {D}\n Answer:
9.To unlock the hidden treasure in {ctx}, you need to choose the correct key.
Which option will open the treasure chest?
A. {A} B. {B} C. {C} D. {D}\n You may choose from 'A', 'B', 'C', 'D'.\n Answer:
10.{ctx}\nQuestion: Reflecting on the emotional stakes and the structure of the narrative, 
which conclusion feels the most genuine?
\n A. {A}\nB. {B}\nC. {C}\nD. {D}\n Answer:
\end{lstlisting}
}
\end{tcolorbox}
\begin{tcolorbox}[title={\textbf{\small Test Prompt of PIQA}},
colback=whitesmoke, colframe=darkblue, , boxrule=2pt, arc=0mm]
{\scriptsize
\begin{lstlisting}[style=mystyle]
1.Solve the following single-choice question by using your common sense reasoning skills.
Choose the correct option and reply with the corresponding letter.
\nQuestion: {goal}\nA. {sol1}\nB. {sol2}\nAnswer:
2.For the situation described by {goal}, consider which solution aligns more closely with how things 
usually work in real life: A. {sol1}\nB. {sol2}. Use logical reasoning to guide your choice. Answer:
3.Given the context of the question, choose the answer that demonstrates the best common 
sense reasoning: {goal}\nA. {sol1}\nB. {sol2}\n Answer format: A/B \nAnswer:
4.In considering the aim set forth in {goal}, visualize the potential consequences of each action 
as if you were directly involved. This visualization can help you identify the better choice:\n
Question: {goal}\nA. {sol1}\nB. {sol2}\nAnswer:
5.Which solution fits the goal based on common sense?
{goal}\n A. {sol1}\nB. {sol2}\n Answer format: A/B \nAnswer:
6.Analyze the following scenario and select the answer that reflects logical reasoning: {goal}
\nA. {sol1}\nB. {sol2}\n Answer format: A/B \nAnswer:
7.Identify the most logical outcome for the situation described: {goal} A. {sol1} B. {sol2} 
Answer format: A/B Remember, the trick is to apply your general knowledge to the scenario. Answer:
8.According to common reasoning, which of the following is the best way to {goal}?
\nA. {sol1}\nB. {sol2}\nAnswer:
9.Which solution best fits the goal based on your general knowledge? {goal}
\n A. {sol1}\nB. {sol2}\n Answer format: A/B \nAnswer:
10.You are about to answer a question that relies on your understanding of basic logic.
Please respond with A or B to indicate your choice.
\nQuestion: {goal}\nA. {sol1}\nB. {sol2}\nAnswer:
\end{lstlisting}
}
\end{tcolorbox}
\begin{tcolorbox}[title={\textbf{\small Test Prompt of Winogrande}},
colback=whitesmoke, colframe=darkblue, , boxrule=2pt, arc=0mm]
{\scriptsize
\begin{lstlisting}[style=mystyle]
1.In the context of {prompt}, which word best completes the sentence? 
Choose: A. {only_option1}. B. {only_option2}.\nAnswer:.
2.When analyzing {prompt}, think about the overall theme. What fits best? 
A. {only_option1}. B. {only_option2}.\nAnswer:.
3.For {prompt}, consider the emotional tone. Which option resonates more?
A. {only_option1}. B. {only_option2}.\nAnswer:.
4.Reflect on {prompt}. Which word logically fills the gap?
A. {only_option1}. B. {only_option2}.\nAnswer:.
5.In {prompt}, which choice aligns with the preceding ideas?
A. {only_option1}. B. {only_option2}.\nAnswer:.
6.When faced with {prompt}, think about the context. What completes it best?
A. {only_option1}. B. {only_option2}.\nAnswer:.
7.For {prompt}, identify the word that maintains the flow of the sentence.
Choose: A. {only_option1}. B. {only_option2}.\nAnswer:.
8.In the case of {prompt}, which option best conveys the intended meaning?
A. {only_option1}. B. {only_option2}.\nAnswer:.
9.Analyze {prompt} for clues. Which word fits the context?
A. {only_option1}. B. {only_option2}.\nAnswer:.
10.When considering {prompt}, which option enhances the clarity of the statement? 
A. {only_option1}. B. {only_option2}.\nAnswer:.
\end{lstlisting}
}
\end{tcolorbox}
\begin{tcolorbox}[title={\textbf{\small Test Prompt of RACE}},
colback=whitesmoke, colframe=darkblue, , boxrule=2pt, arc=0mm]
{\scriptsize
\begin{lstlisting}[style=mystyle]
1.After reading the article, analyze the question and choose the best answer
based on the details and themes discussed. Look for clues within the text that
align with one of the options.\nArticle:\n{article}\n\nQuestion:
{question}\nOptions: \nA. {A}\nB. {B}\nC. {C}\nD. {D}\nAnswer:
2.Article:\n{article}\nAfter reading the passage, please answer the following question:
\n{question}\nA. {A}\nB. {B}\nC. {C}\nD. {D} \nAnswer:
3.Carefully read the following article and answer the question by selecting the correct option.
Respond with A, B, C, or D.\n\nArticle:\n{article}\n\n
Q: {question}\n\nA. {A}\nB. {B}\nC. {C}\nD. {D}\nAnswer:
4.Read the text carefully and answer the question by choosing the most appropriate option.
Evaluate the relevance of each choice to the main points discussed.
\nArticle:\n{article}\n\nQuestion: {question}\nOptions: \nA. {A}\nB. {B}\nC. {C}\nD. {D}\nAnswer:
5.Describe the setting of the article. 
{question}\n{article}\nA. {A}\nB. {B}\nC. {C}\nD. {D} \nAnswer:
6.While reading the {article}, highlight or make mental notes of significant details. 
The {question} is asking [describe the specific query]. 
Now evaluate the options:\nA. {A}\nB. {B}\nC. {C}\nD. {D} \nAnswer:
7.After carefully analyzing {article}, determine which of the following options best 
answers the question: 
{question}. A. {A}\nB. {B}\nC. {C}\nD. {D} \nAnswer:
8.Read {article} with a focus on answering {question}. Choose the most suitable option. 
Article: {article} Question:{question} Options: A. {A} B. {B} C. {C} D. {D} 
Trick: Be cautious of answer choices that seem too extreme. Your answer is just one letter. Answer:
9.Article:\n{article}\nFrom the information in the article, identify the correct 
answer to the following question: \n{question}\nA. {A}\nB. {B}\nC. {C}\nD. {D} \nAnswer:
10.When {article} mentions {question}, which option best describes the author's attitude?
\nA. {A}\nB. {B}\nC. {C}\nD. {D} \n// Pay attention to the tone of the author.
Look for words that convey emotions or opinion to determine the attitude.Answer:
\end{lstlisting}
}
\end{tcolorbox}
\subsection{Baseline prompt}\label{baseline prompt}
In this section, we present the best prompts generated or filtered using the baseline for training.
\begin{tcolorbox}[title={\textbf{\small Prompt of Hellaswag}},
colback=whitesmoke, colframe=darkblue, , boxrule=2pt, arc=0mm]
{\scriptsize
\begin{lstlisting}[style=mystyle]
TopAccuracy prompt: 
Given the context {ctx}, predict the correct ending by choosing the most logical option.
\n A. {A}\nB. {B}\nC. {C}\nD. {D}\n You may choose from 'A', 'B', 'C', 'D'.\n Answer:

User-specified prompt: 
{ctx}\n Question: {Question}\n A. {A}\nB. {B}\nC. {C}\nD. {D}\n 
You may choose from 'A', 'B', 'C', 'D'.\n Answer:

BATprompt : 
Given the context below, predict the most logical ending by choosing the correct option 
from the provided choices. Ensure your choice aligns with the context and is the most coherent
conclusion. \n Context: {ctx}\n 
Question: Which ending makes the most sense?\n A. {A}\nB. {B}\nC. {C}\nD. {D}\n 
You may choose from 'A', 'B', 'C', 'D'.\n Answer:

ZOPO prompt:
Based on {ctx}, which option is the most likely correct ending?
Consider the overall context, character motivations, and any foreshadowing. 
Trick: Analyze the consistency of each option with the established details. 
A. {A}\nB. {B}\nC. {C}\nD. {D}\n You may choose from 'A', 'B', 'C', 'D'.\n Answer:

\end{lstlisting}
}
\end{tcolorbox}
\begin{tcolorbox}[title={\textbf{\small Prompt of PIQA}},
colback=whitesmoke, colframe=darkblue, , boxrule=2pt, arc=0mm]
{\scriptsize
\begin{lstlisting}[style=mystyle]
TopAccuracy prompt: 
Use both common sense and logical reasoning to determine the correct solution for the goal:
{goal}\n A. {sol1}\nB. {sol2}\n Answer format: A/B \nAnswer:

User-specified prompt: 
There is a single choice question. Answer the question by replying A or B.'\n 
Question: {goal}\nA. {sol1}\nB. {sol2}\nAnswer:

BATprompt : 
You should use both common sense and logical reasoning to determine the most appropriate 
solution for the following goal. Carefully evaluate the provided options and choose the 
one that best aligns with the goal. Goal: {goal}\nA. {sol1}\nB. {sol2}\nAnswer:

ZOPO prompt:
To solve this common sense reasoning question, consider which of the two options seems 
more plausible based on everyday knowledge and logic.
\nQuestion: {goal}\nA. {sol1}\nB. {sol2}\n
Think about the practical implications of each choice to determine the correct answer.\nAnswer:
\end{lstlisting}
}
\end{tcolorbox}
\begin{tcolorbox}[title={\textbf{\small Prompt of Winogrande}},
colback=whitesmoke, colframe=darkblue, , boxrule=2pt, arc=0mm]
{\scriptsize
\begin{lstlisting}[style=mystyle]
TopAccuracy prompt: 
Question: {prompt}\nA. {only_option1}\nB. {only_option2}\nAnswer:

User-specified prompt: 
There is a single choice question,  you need to choose the correct option to fill in the blank. 
Answer the question by replying A or B.\n 
Question:{prompt}\nA. {only_option1}\nB. {only_option2}\nAnswer:

BATprompt : 
Complete the following sentence by selecting the most contextually appropriate option. 
Carefully consider the meaning and context of the sentence to make your choice. 
Question: {prompt}\nA. {only_option1}\nB. {only_option2}\nAnswer:

ZOPO prompt:
Question: Choose the correct modal verb: {prompt}\nA. {only_option1}\nB. {only_option2}\nAnswer:.
\end{lstlisting}
}
\end{tcolorbox}
\begin{tcolorbox}[title={\textbf{\small Prompt of RACE}},
colback=whitesmoke, colframe=darkblue, , boxrule=2pt, arc=0mm]
{\scriptsize
\begin{lstlisting}[style=mystyle]
TopAccuracy prompt: 
Read the following article carefully: {article}. After reading, answer the question: {question}. 
Choose the correct option from the choices provided: 
\nA. {A}\nB. {B}\nC. {C}\nD. {D} \n 
Trick: Focus on the main idea and supporting details in the article. 
Output: Only the letter of the correct answer.\nAnswer:

User-specified prompt: 
Article:\n{article}\nQuestion:\n{question}\nA. {A}\nB. {B}\nC. {C}\nD. {D} \nAnswer:

BATprompt : 
Please read the passage carefully, focusing on the main ideas and supporting details. 
Answer the question that follows by choosing the best option from the choices provided. 
Ensure your response is based solely on the information in the passage. Output only the 
letter of the correct answer. Article:\n{article}
\nQuestion:\n{question}\nA. {A}\nB. {B}\nC. {C}\nD. {D} \nAnswer:

ZOPO prompt:
A reading comprehension question is before you. Read the article and answer the question 
by selecting A, B, C, or D.\n\nArticle:\n{article}\n\n
Q: {question}\n\nA. {A}\nB. {B}\nC. {C}\nD. {D}\nAnswer: 
\end{lstlisting}
}
\end{tcolorbox}
\end{document}